\documentclass[sigconf]{acmart} 
\usepackage{microtype}
\usepackage{xcolor}
\usepackage{xspace}
\usepackage{hyperref}
\usepackage{enumitem}
\usepackage{cleveref}
\usepackage{acmart-taps}

\definecolor{linkColor}{RGB}{71,115,197}
\definecolor{mypurple1}{RGB}{206, 50, 217}
\definecolor{myblue1}{RGB}{0,0,255}
\definecolor{myorange}{RGB}{200, 128, 0}

\definecolor{queryBlue}{RGB}{118,169,250}
\definecolor{keyRed}{RGB}{249, 128, 128}
\definecolor{valueGreen}{RGB}{49, 196, 141}

\definecolor{TEColorscale}{RGB}{114,106,208}
\definecolor{BPColorscale}{RGB}{228,132,45}
\definecolor{VDColorscale}{RGB}{76,190,154}

\definecolor{HighColorscale}{RGB}{255,127,14}
\definecolor{LowColorscale}{RGB}{31,119,180}

\definecolor{exploreColorscale}{RGB}{63,131,248}
\definecolor{readerColorscale}{RGB}{227,160,8}
\definecolor{generatorColorscale}{RGB}{249,128,128}
\definecolor{casualColorScale}{RGB}{156, 163, 175}

\newcommand{\tf}[0]{Transformer\xspace}
\newcommand{\tfs}[0]{Transformers\xspace}
\newcommand{\tool}[0]{Transformer Explainer}

\AtBeginDocument{%
  }

\copyrightyear{2026}
\acmYear{2026}
\setcopyright{cc}
\setcctype{by}
\acmConference[CHI '26]{Proceedings of the 2026 CHI Conference on Human Factors in Computing Systems}{April 13--17, 2026}{Barcelona, Spain}
\acmBooktitle{Proceedings of the 2026 CHI Conference on Human Factors in Computing Systems (CHI '26), April 13--17, 2026, Barcelona, Spain}
\acmPrice{}
\acmDOI{10.1145/3772318.3791725}
\acmISBN{979-8-4007-2278-3/2026/04}

\begin{document}

\title{\tool{}: Learning LLM Transformers with  Interactive Visual Explanation and Experimentation}

\author{Aeree Cho}
\orcid{0009-0002-5135-8136}
\affiliation{%
% \institution{College of Computing}
  \institution{Georgia Tech}
  \city{Atlanta}
  % \state{Georgia}
  \country{USA}}
\email{aeree@gatech.edu}

\author{Grace C. Kim}
\orcid{0009-0004-3845-9463}
\affiliation{%
% \institution{College of Computing}
  \institution{Georgia Tech}
  \city{Atlanta}
  % \state{Georgia}
  \country{USA}}
\email{gracekim3@gatech.edu}

\author{Alexander Karpekov}
\orcid{0009-0006-5317-2991}
\affiliation{%
% \institution{College of Computing}
  \institution{Georgia Tech}
  \city{Atlanta}
  % \state{Georgia}
  \country{USA}}
\email{alex.karpekov@gatech.edu}

\author{Seongmin Lee}
\orcid{0000-0002-1950-5004}

\affiliation{%
% \institution{College of Computing}
  \institution{Georgia Tech}
  \city{Atlanta}
  % \state{Georgia}
  \country{USA}}
\email{seongmin@gatech.edu}

\author{Alec Helbling}
\orcid{0009-0007-8846-6460}
\affiliation{%
% \institution{College of Computing}
  \institution{Georgia Tech}
  \city{Atlanta}
  % \state{Georgia}
  \country{USA}}
\email{alechelbling@gatech.edu}

\author{Benjamin Hoover}
\orcid{0000-0001-5218-3185}
\affiliation{%
    \institution{IBM Research AI}
  \city{Cambridge}
  % \state{Massachusetts}
  \country{USA}
  % \institution{College of Computing}
  \institution{Georgia Tech}
  \city{Atlanta}
  % \state{Georgia}
  \country{USA}}
\email{bhoov@gatech.edu}

\author{Zijie J. Wang}
\orcid{0000-0003-4360-1423}
\affiliation{%
% \institution{College of Computing}
  \institution{Georgia Tech}
  \city{Atlanta}
  % \state{Georgia}
  \country{USA}}
\email{jayw@gatech.edu}

\author{Minsuk Kahng}
\authornote{Corresponding author}
\orcid{0000-0002-0291-6026}
\affiliation{%
% \institution{Department of Computer Science and Engineering} 
  \institution{Yonsei University}
  \city{Seoul}
  \country{Republic of Korea}}
\email{minsuk@yonsei.ac.kr}

\author{Duen Horng (Polo) Chau}
\authornotemark[1]
\orcid{0000-0001-9824-3323}
\affiliation{%
% \institution{College of Computing}
  \institution{Georgia Tech}
  \city{Atlanta}
  % \state{Georgia}
  \country{USA}}
\email{polo@gatech.edu}

\renewcommand{\shortauthors}{Cho et al.}

\begin{abstract}
The Transformer architecture underpins modern large language models powering state-of-the-art text generation and AI applications. However, its complexity makes it difficult for non-experts to learn. Existing resources often lack interactivity, rely on static descriptions of simplified architectures, or fail to reflect models’ behavior with real data. To address this gap, we introduce Transformer Explainer, an interactive visualization tool for non-experts to learn Transformers. The tool integrates an overview illustrating the Transformer's data flow with on-demand explanations that gradually reveal mathematical details. Smooth transitions across abstraction levels highlight the interplay between high-level structures and low-level operations. Running a live GPT-2 instance directly in the browser, Transformer Explainer empowers learners to experiment with custom input and hyperparameters without setup, observing next-token predictions in real time. A 90-participant user study showed that our tool offered significant advantages in
improving user understanding and engagement. Transformer Explainer has attracted over 490,000 users.
\end{abstract}



\begin{teaserfigure}
  \includegraphics[width=\textwidth]{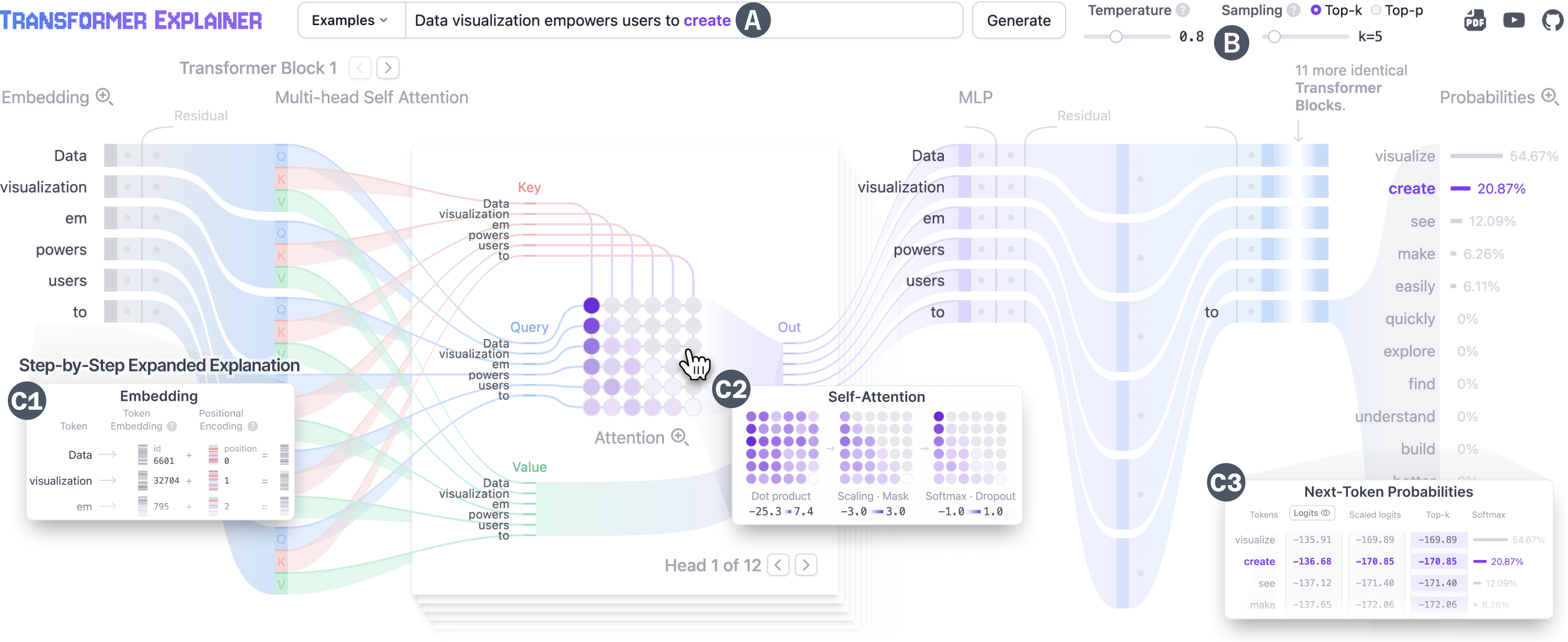}
  \caption{
\tool{} helps users \textbf{(A)} visually explore how a Transformer text-generation model (GPT-2) processes input text into a prediction for the next token, \textbf{(B)} interactively manipulate often-confused hyperparameters, such as temperature and sampling strategies, to understand their effects on prediction determinism;
  and \textbf{(C)} seamlessly transition between abstraction levels to visualize the interplay between high-level model structures and low-level mathematical operations for \textbf{(C1)} embedding, \textbf{(C2)} self-attention, and \textbf{(C3)} next-token probabilities.
  }
    \Description{Figure that provides an overview of the Transformer Explainer interface for visualizing how a Transformer text-generation model processes input tokens into next-token predictions. The diagram shows input words such as “Data,” “visualization,” and “powers” being converted into embeddings, which then pass through multi-head self-attention and MLP layers inside repeated Transformer blocks. The visualization highlights attention patterns, where query, key, and value connections are shown linking tokens. On the right side, output probabilities are displayed, with the token “create” assigned a probability of 20.87\%. The interface also includes controls for temperature and sampling strategies, such as top-k and top-p, allowing users to observe how these hyperparameters influence prediction determinism and randomness. Insets provide step-by-step expanded explanations of embeddings, attention, and next-token probability calculations, illustrating the relationship between high-level model structure and detailed mathematical operations.}
  \label{fig:teaser}
\end{teaserfigure}

\maketitle

\section{Introduction}
\label{sec:intro}

The \tf{} \cite{10.5555/3295222.3295349} has become the state-of-the-art neural network architecture across diverse domains, including natural language processing and computer vision, and forms the backbone of large language models (LLMs) such as \href{https://openai.com/chatgpt}{\textcolor{linkColor}{ChatGPT}}, \href{https://www.deepseek.com/}{\textcolor{linkColor}{DeepSeek}}, and \href{https://gemini.google.com/app}{\textcolor{linkColor}{Gemini}}.
However, its complex internal structure poses significant learning challenges for non-experts, hindering their understanding and engagement \cite{rogers-etal-2020-primer}.
Existing resources typically rely on static or non-interactive explanations
that do not support experiential learning \cite{alammar, yt_3blue1brown}  %
and may not fully reflect the model's behavior with real data \cite{llm3d}.
Research from our HCI community has demonstrated the significant benefits of interactive visual explanations in empowering learners to more easily engage with and learn complex concepts \cite{Kehoe2001265, fouh2012role, CNNExplainer, kahng2019gan, tensorflowplayground}.
For non-experts wishing to take their first steps toward understanding this technology, engaging and interactive explanations may be crucial for supporting such learning.

\textbf{Key challenges in designing learning tools for \tfs{}.}
An increasing body of research is leveraging interactive visualizations to explain deep learning concepts.
However, \tfs{} introduce unique challenges that require novel visual designs.
For example, unlike earlier models such as Convolutional Neural Networks (CNNs),
where operations can be made easier to understand through visualizations (e.g., filters sliding over input images \cite{CNNExplainer}), \tfs{} operate on high-dimensional numerical representations of text that may be harder to understand.
The mathematical operations performed on these representations add further complexity: 
\tfs{} consist of \textit{many repeating blocks}, each containing many interacting operations (\autoref{fig:diagram})---requiring learners to develop a mental model of not only how individual components work but also how they interact \cite{ferrando2024information, lu2021influence}.
Notably, the \textit{multi-head self-attention} mechanism, unique to \tfs{} and present in every block \cite{10.5555/3295222.3295349}, is significantly more complex than other model operations like CNN convolutions \cite{rogers-etal-2020-primer}, 
as it involves multiple matrix operations that enable every input text \textit{token} (typically a word, subword, or character; \autoref{sec:background}) to simultaneously interact with \textit{every other token}.
Furthermore, \tf{} prediction is \textit{autoregressive} \cite{mann2020language}, which introduces additional challenges---each output token depends on all previously generated tokens, with key sampling hyperparameters influencing every step of the generation process.
Therefore, the difficulty in learning about \tfs{} originates not only in the complexity of each component, but also in understanding the dynamic, high-dimensional, iterative, and probabilistic interplay between the large number of components.
Existing learning resources have not adequately addressed such crucial learning needs, as they often rely on static descriptions of simplified architectures, lack real-time interactivity, or may not fully reflect the models’ behavior with real data.
We aim to bridge this critical gap.

In this work, we contribute:
\begin{enumerate}[itemsep=1mm, topsep=1mm, parsep=0mm, leftmargin=4mm]
    \item \textbf{\tool{}, an interactive tool for non-experts to learn and experiment with Transformers for text generation} (\autoref{fig:teaser}). \tool{} overcomes unique challenges in developing interactive learning tools for Transformers (\autoref{sec:design_challenges}), distilled through a literature review of visual learning resources and analytics tools for Transformers, and more broadly, deep learning (\autoref{sec:related_work}). Through iterative design and consultation with machine learning instructors (\autoref{sec:iteration}), Transformer Explainer offers statistically significant advantages over conventional learning resources (blog and video) in improving non-experts’ understanding of Transformer concepts (\autoref{sec:findings}).

    \item \textbf{Novel techniques and interactive system designs} to improve non-expert users' understanding of complex Transformer concepts.
    \begin{itemize}
    \item \tool{} adopts a token-centric flow-based visual design, which is supported by recent studies that highlight the importance of tracing information flow through a \tf{} model to understand its behavior \cite{ferrando2024information, lu2021influence}. 
    The flow-based design offers an overview that visually communicates the \textit{token embedding data flow} across the model components, illustrating how \textit{inputs} are processed and transformed across model components to reach the final output: the \textit{next token} (\autoref{sec:overview}).
    
    \item    
    Our tool enables users to interactively expand the model components through animated transitions that preserve context and data flow, to explore \textit{step-by-step} explanations of
    mathematical details (\autoref{fig:teaser}:~C2).
    By smoothly transitioning between abstraction levels, users can see the interplay between low-level, detailed mathematical operations and high-level model structures, gaining a comprehensive understanding of \tfs{} (\autoref{sec:expandedView}).
    
    \item 
    Our tool allows users to input their own text (\autoref{fig:teaser}A) and directly manipulate sampling hyperparameters (\autoref{fig:teaser}B), enabling them to observe model behavior under different conditions in real time. Unlike existing educational resources—which often overlook how the generated probability distribution determines next-token predictions and how hyperparameters shape output variability \cite{llm3d, alammar}—our interface visualizes these effects (\autoref{fig:teaser}:~C3). For instance, while temperature is frequently anthropomorphized as a “creativity” control, users can experiment to see how it actually modifies the probability distribution and randomness of next-token predictions (\autoref{sec:generation}).

    \item 
    We introduce a guided learning feature---an interactive, step-by-step text explanation card embedded within the tool (\autoref{sec:guided-learning}). 
    Users can progressively learn Transformer concepts while linking them to visual components and interactive actions (e.g., adjusting hyperparameters and highlighting the resulting changes). 
    Unlike conventional onboarding tutorials \cite{stoiber2023visualization}, guided learning supports both conceptual understanding and tool usage, serving as an in-situ explanation that can be accessed whenever additional clarification is needed.
    \end{itemize}

    \item \textbf{Design lessons derived from a user study
    } on interactive visual explanation and experimentation.
    To evaluate \tool{}'s usability and effectiveness, we conducted a 90-participant between-subjects user study, 
    which identified key advantages of the tool and confirmed its effectiveness in helping non-experts better understand complex model architectures and underlying operations (\autoref{sec:user_study}). 
    Our findings highlight important lessons for future AI education tools, showing how interactivity enhances understanding, flow-based visualization clarifies complex architectures, and guided learning reduces entry barriers.
    \item \textbf{Open-sourced, web-based implementation powered by a live model} that broadens public  access to our tool.
    Unlike many existing tools that require specialized software setups or lack inference functionality~\cite{tf-viz-survey}, \tool{} hosts a live, fully-functional \tf{} model that runs directly in the user's web browser.
    We selected the GPT-2 model for its widespread recognition, fast inference speed, and architectural similarities to more advanced models such as GPT-4 or later, making it suitable for educational use.
    Anyone can access \tool{} directly in their browser without the need for installation or specialized hardware, allowing a large number of users to explore and learn from the tool simultaneously on their own devices.
    Our tool is open-source; a demo video, source code, and a live demo are included as supplementary material.
    Since its launch, \tool{} has reached over 490,000 users across more than 200 countries and continues to contribute to the democratization of modern generative AI education.

\end{enumerate}
\section{Background for \tfs{}}
\label{sec:background}
\begin{figure}[H]
\centering
\includegraphics[width=\linewidth]{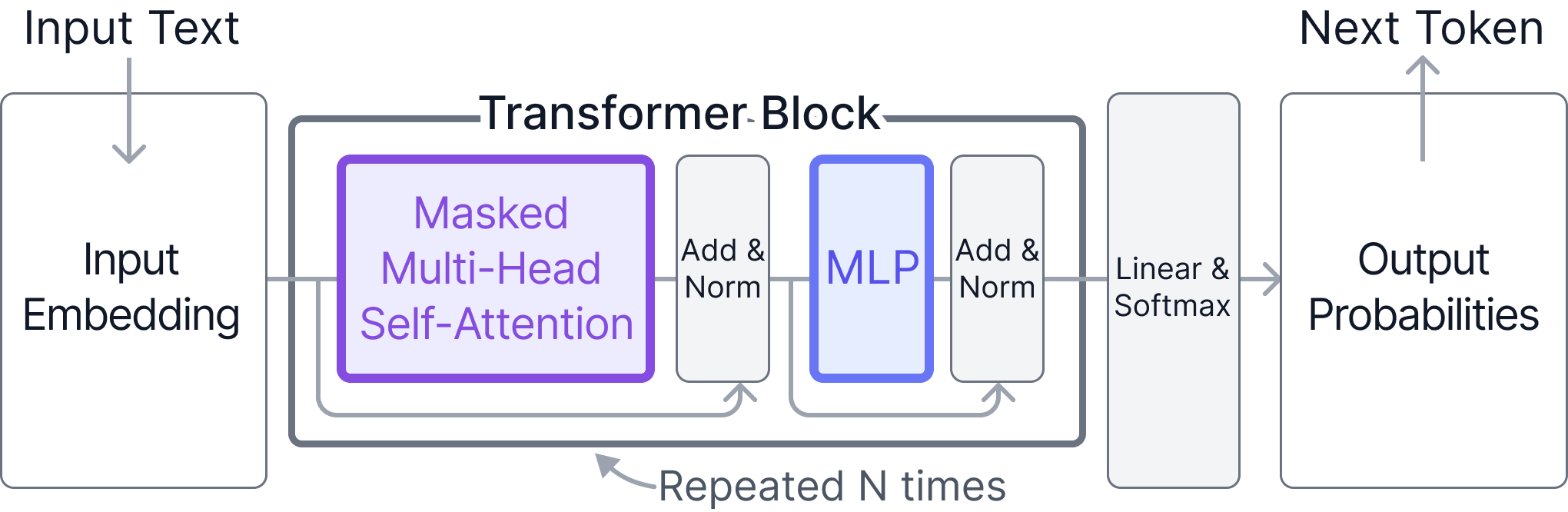}
\caption{
Diagram illustrating the data flow of the \tf{} architecture (decoder-only), 
which consists of many components:
input text is converted to input embeddings that pass through repeated identical \tf{} blocks, 
each containing masked multi-head self-attention mechanisms, followed by multi-layer perceptron (MLP) layers with residual connections (Add) and normalization (Norm). 
The final linear and softmax layers output probabilities for the next-token prediction.
} 
\label{fig:diagram}
\Description{Figure that provides a schematic overview of the decoder-only Transformer architecture. The diagram begins with input text converted into input embeddings, which are then passed through a sequence of repeated Transformer blocks. Each block contains masked multi-head self-attention and multi-layer perceptron (MLP) layers, along with residual connections and normalization steps. After flowing through multiple stacked blocks, the outputs proceed to a linear transformation and softmax layer, which together generate the probability distribution over the next token. The illustration emphasizes the repeated structure of Transformer blocks and how embeddings are progressively transformed into output probabilities for text generation.}
\end{figure}

In this section, we provide a high-level overview of a \tf{} model architecture \cite{10.5555/3295222.3295349} (\autoref{fig:diagram}) in the context of text generation, which will help ground our discussion in the paper. 
\tfs{} like GPT-2 that are used to purely generate text are known as \textit{decoder-only} \tfs{}, and they contain all the core components of the original \tf{}~\cite{10.5555/3295222.3295349}. 
Because \tool{} focuses on explaining decoder-only \tfs{}, we use the term \textit{\tf{}} to specifically refer to decoder-only variants throughout this work.

To generate text, a \tf{} performs the following processes:
First, the input text is split into \textit{tokens}: units of text that are typically words, subwords, or characters 
(e.g., \texttt{``Data visualization empowers''}~$\rightarrow$~\texttt{[``Data'', `` visualization'', `` em'', ``powers'']}).
Each token is converted to a \textit{token embedding} (i.e., its numerical vector representation), and positional encoding is added to preserve the order of tokens. 
This embedding passes through multiple \tf{} blocks, each containing a \textit{self-attention} mechanism with causal masking to prevent tokens from attending to future tokens in the sequence. 
Self-attention transforms each token embedding in a sequence of embeddings into a ``query'', ``key'', and ``value'', and the relationship between these three representations is best understood through an intuitive analogy: 
to predict the next token, the current token's \textit{query} identifies which of the preceding tokens hold the most relevant information---specifically, the \textit{values} associated with the most similar \textit{keys}. 
The degree of similarity between a query and a key is quantified
as the \textit{attention score}.
A single self-attention mechanism is known as an attention \textit{head}. 
Stacking individual heads in parallel forms \textit{multi-head self-attention},\footnote{Original architecture \cite{10.5555/3295222.3295349} proposed 6 \tf{} blocks, and 8 heads.} which enables the model to form a richer contextual understanding of the input, where each head can focus on different aspects of the input. 

After the self-attention mechanism, the transformed embedding passes through an \textit{MLP}, or multi-layer perceptron, which further increases the representational capacity of the \tf{}. 
Across the multiple \tf{} blocks, earlier blocks tend to capture low-level features, while later blocks represent more abstract semantic features.
Finally, the model projects the transformed embedding into a probability distribution over all possible tokens, determining the likelihood of each token being the next in the sequence.
During inference, sampling hyperparameters such as \textit{temperature} control the sharpness or smoothness of the probability distribution, while sampling strategies such as \textit{top-k} or \textit{top-p} are used to select the next-token candidates.
This process continues iteratively until the model fully generates a sequence of text.

\section{Related Work}
\label{sec:related_work}

\subsection{Traditional Visual Learning Resources for \tfs{}}
Most introductory resources for \tf{}-based models have been offered as blog posts or video tutorials. A well-known example is The Illustrated \tf{} by Jay Alammar \cite{alammar}, which provides static illustrations with explanatory text. 
Video tutorials are also popular, with some focusing on theoretical concepts \cite{yt_3blue1brown} and others on code implementation \cite{yt_karpathy}. 
While these resources reach a large audience, their static and linear formats often struggle to convey the input-dependent dynamic processes in \tfs{}, making it difficult for learners to understand them in detail and stay engaged in the learning process \cite{fouh2012role}.
This challenge suggests a need for learning materials that support interactivity and real-time feedback.

\subsection{Interactive Articles for Deep Learning Education}
Beyond static blogs, interactive “explorable explanations”~\cite{victor} emerged as a new medium for deep learning education. 
Following Chris Olah's interactive blog posts in 2014~\cite{olah2014neural}, interactive articles with visualizations \cite{you2024pandapandaunderstandingadversarial, r2d3aintroml, chen2021empiricalneural, conlen2024dimred} began gaining traction \cite{CNNExplainer}.
Distill, a scientific journal for machine learning, was also established to publish articles featuring interactive graphics and explorable explanations \cite{görtler2019gaussian, goh2017why}.
While such articles provide more engaging learning experiences than static blogs, their interactivity remains limited: users typically follow predefined storylines and cannot freely manipulate inputs or explore alternative scenarios. 
Learners often need to scroll back and forth between overviews and detailed explanations, which disrupts continuity and makes it difficult to connect concepts across levels of detail.

\subsection{Interactive Visual Tools for Deep Learning Education}
\label{sec:interactive_tools_for_dl_education}
More recently, to address the limitations of the traditional educational mediums mentioned above, fully interactive visualization tools have been developed for deep learning education.
Early systems like ConvNetJS MNIST Demo and TensorFlow Playground offered intuitive parameter tweaking at the level of small models \cite{karpathyConvNetJSMNISTDemo2016, tensorflowplayground}. 
Later, tools such as GAN Lab and CNN Explainer advanced the genre by introducing interactive graphics on real or synthetic data, bridging low-level mathematical details with high-level conceptual explanations \cite{kahng2019gan, CNNExplainer}. 

Recent work has proposed diverse ``explainer'' systems, but these do not converge on a single, settled design pattern. Instead, different model families and application contexts demand distinct visualization and interaction techniques.
For example, to explain diffusion models, Diffusion Explainer lets users change various model parameters and observe their effects through timestep-based animations \cite{lee2024diffusion} and Patch Explorer visualizes internal representations as an interactive 3D heatmap \cite{patch}.
GNN101 overlays a graph neural network as stacked visual layers and connects nodes across layers to depict layer-wise message passing \cite{gnn}.
Raise Playground uses a block-based programming interface to let learners practice and learn AI concepts hands-on \cite{raisePlayground}.
% Recently, Diffusion Explainer extended this approach to help non-experts more easily learn  modern generative text-to-image architectures  \cite{lee2024diffusion}. 
% After our tool's release, VAE Explainer built on our design to explain 
% variational autoencoders \cite{bertucci2024vae}.
PromptAid helps non-experts and practitioners explore, perturb, and test prompts for large language models with dashboard-style collection of visualizations \cite{mishra2025promptaid}.

To our knowledge, the only existing interactive visualization tools designed for learning about \tfs{} are LLM Visualization \cite{llm3d} and TransforLearn \cite{gao2023transforlearn}.
LLM Visualization offers a step-by-step guide through the \tf{} architecture visualized in 3D, 
but it lacks support for custom user inputs 
and focuses on presenting mathematical details 
without abstraction levels, which may overwhelm beginners.
TransforLearn
supports interactive model exploration for machine translation but has key limitations that affect its educational impact. Its high-level overview displays heatmaps of embedding vectors, but otherwise relies on static text and diagrams that do not adapt to input changes,
hindering user understanding \cite{fouh2012role}.
Users can click on individual components to view separate visualizations, but the absence of continuous data flow makes it difficult to track how data transforms across the architecture. Moreover, the emphasis on embeddings in the overview may detract from learning self-attention, the core \tf{} component ~\cite{NIU202148, 10.5555/3295222.3295349}. Finally, TransforLearn requires a server for machine translation tasks and is not deployed online.
\tool{} overcomes all the above limitations as the only online interactive visualization tool for learning \tfs{}, offering real-time inference with custom user input.

\subsection{Visual Analytics Tools for Transformer Interpretability}
While educational tools are designed to help non-experts learn about Transformer concepts, another large body of work focuses on visual analytics systems designed for researchers and practitioners to interpret and analyze the internal behaviors and computations of Transformer models. These systems typically target expert users, offering fine-grained inspection of attention patterns, hidden states, or neuron activations.
A prominent line of work investigates attention mechanisms. AttentionViz provides global overviews of attention patterns across layers and heads in both language and vision Transformers \cite{yeh2023attentionviz}. Attention Flows introduces mechanisms to query, trace, and compare attention shifts during pretraining and fine-tuning \cite{derose2020attention}. Other systems extend this approach to specific domains, such as head-level analysis for Vision Transformers (ViTs) \cite{li2023does}, cross-modal interactions in vision-language Transformers \cite{aflalo2022vl}, and examines reasoning in Transformer-based Visual Question Answering (VQA) models \cite{jaunet2021visqa}.

Beyond attention, attribution-based systems offer complementary interpretability. VEQA analyzes open-domain QA with BERT by visualizing retrieval–reader decision flows \cite{shao2023visual}. Recent neuron-level methods further attribute knowledge in LLMs by identifying value and query neurons \cite{yu2023neuron}. The family of Logit Lens approaches \cite{logitlens, tunedlens, attributelens, futurelens} performs layer-wise decoding to show how next-token predictions evolve across intermediate layers, providing an alternative way to interpret the information encoded internally by Transformers.
Additionally, work from Anthropic has advanced interpretability through interactive, explorable articles that enable users to analyze model components \cite{elhage2021mathematical, ameisen2025circuit, lindsey2025biology} and with tools like Circuit Tracer \cite{circuit-tracer}, which enables users to trace and visualize circuits formed by interactions between model components that contribute to specific model behaviors.
These tools provide powerful interpretability for expert analysis but are not intended for non-experts, as they often require deep ML background knowledge to use effectively.
Our goal, by contrast, is to help non-experts acquire this foundational background by introducing the core architectural concepts and data-flow processes that underlie Transformer models.

\subsection{How Our Work Fills Unique Research Gaps}
In summary, prior work for Deep Learning education can be grouped into (1) traditional static resources, such as blogs and videos; (2) interactive articles that guide readers through predefined narratives;
(3) interactive educational visualization tools that support non-experts but do not address Transformers, while those that do are limited (\autoref{sec:interactive_tools_for_dl_education});
and (4) visual analytics systems for experts that provide in-depth interpretability but are not designed for beginners.

Transformer Explainer fills a unique gap between these categories. Unlike traditional or narrative-driven media, it provides real-time, interactive feedback on custom user input. 
Unlike existing educational tools for Transformers, it supports fully interactive, real-time exploration and complete data flow visualization, using levels of abstraction to introduce the entire model architecture while gradually revealing details to avoid overwhelming users.
And unlike expert interpretability systems, it is designed to support non-expert learners, aiming to make complex architectures comprehensible. By combining these strengths, Transformer Explainer advances the landscape of educational resources for deep learning, offering a novel, fully online interactive system for learning Transformers at scale. %
\section{Design Challenges}
\label{sec:design_challenges}

Transformer-based LLMs introduce fundamentally different challenges across structural, visual, and computational dimensions that existing explainer systems  designed for vision models or classification models \cite{CNNExplainer, kahng2019gan, tensorflowplayground, karpathyConvNetJSMNISTDemo2016, lee2024diffusion} did not address.
To design \tool{}, we identified four main design challenges (C1-C4) related to \tf{} models:

\aptLtoX{\begin{enumerate}[label=\textbf{C\arabic*.}, leftmargin=*, ref=\textbf{C\arabic*}]
    \item[C1.] \textbf{Understanding How Input Text Is Processed Across Complex Model Structures.}\label{challenge:flow}
\tfs{} are complex models composed of many components, such as multi-head self-attention and multi-layer perceptron (MLP), repeated across multiple layers \cite{10.5555/3295222.3295349, ganesh2021compressing} (\autoref{fig:diagram}).
While existing resources have attempted to provide an overview of the model \cite{alammar}, 
they either present all details at once \cite{llm3d}, which may overwhelm beginners, or display model components in disconnected views, often using disjoint visual encodings that increase users’ cognitive load \cite{tufte1983visual}.
However, research shows that presenting  all layer operations and their connections in a unified view has the potential to help users better follow how input data (i.e., token embedding) is transformed into final predictions.
Such attempts typically use diagram-style layouts that place each layer or component as a boxed node and connect them with a line, which works reasonably well for vision models because each node can be visualized as an image \cite{CNNExplainer, kahng2019gan, karpathyConvNetJSMNISTDemo2016, lee2024diffusion}. 
In contrast, Transformers operate on token representations and do not follow the relatively simple feedforward structure of classification models \cite{CNNExplainer, tensorflowplayground, karpathyConvNetJSMNISTDemo2016}. As token representations are interactively transformed as they go through attention, visualizing their transformations introduces unique challenges.
There have been attempts to apply similar diagram-style approaches to Transformers, using component-level boxes connected by simple lines \cite{gao2023transforlearn}, but, they fail to foreground tokens as the primary and coherent visual element and cannot convey an end-to-end transformation path of how each token representation evolves throughout the model.
Hence, an innovative visualization is needed to create a visual summary of the \tf{} that maintains a connected view and preserves data flow.
Adopting an information-flow perspective has potential to help non-experts conceptually link inputs, intermediate computations, and outputs into a coherent process \cite{ferrando2024information}.

\item[C2.] \textbf{Mathematical Complexity in Multi-Head Self-Attention.}
\label{challenge:math}
Non-experts often struggle to understand the underlying operations in deep learning models \cite{tensorflowplayground}.
In models like CNNs for images, operations can be more easily understood via visual metaphors---such as filters sliding over input images \cite{CNNExplainer}.
In contrast, \tfs{} operate on high-dimensional numerical representations of text, which are less interpretable on their own \cite{NIU202148}.
The complexity deepens with operations like self-attention, where every token interacts with every other token, and with attention-specific operations such as projecting embeddings into Q, K, and V vectors and partitioning them into multiple heads \cite{10.5555/3295222.3295349, Baan2019UnderstandingMA}.
This multi-head attention is foundational to how the model selects, gathers, and combines relevant information from different perspectives \cite{10.5555/3295222.3295349}.
Understanding multi-head attention therefore helps learners understand the concrete mechanism in which words are used to predict the next token, demystifying how the model uses context. 
% 
% What looks like a Transformer “understanding context” actually arises from each token using multiple attention heads to select, gather, and combine relevant information from different perspectives. Understanding multi-head attention therefore helps learners explain which words the model relied on to predict the next token, and why it sometimes misses key context or forms spurious connections, in terms of concrete mechanisms rather than treating the model as a black box.
% 
Existing resources typically explain these operations separately and in detail, but do not effectively help non-experts visually connect how the components work together to transform the token representations \cite{alammar, gao2023transforlearn}. 
Moreover, detailed mathematical explanations are often foregrounded, which can overwhelm non-expert learners with limited mathematical background and discourage further engagement. This gap highlights the need for more accessible explanation, motivating novel visualization and interaction techniques that can unpack and clarify these operations.

\item[C3.] \textbf{Understanding Hyperparameters' Impact on Prediction Variability.} 
\label{challenge:parameters}
\tf{} models generate a probability distribution over possible tokens, from which the next token is sampled during autoregressive generation.
Yet many educational resources either fix the prompt and the output, or they omit how the distribution is constructed and how sampling hyperparameters (e.g., temperature, top-$k$, top-$p$) reshape both the candidate set and the final choice \cite{llm3d, gao2023transforlearn}.
As a result, many learners remain unaware of these underlying mechanisms, often viewing \tfs{} as magical or even anthropomorphizing them \cite{stochasticparrots}.
% This highlights a key research challenge: 
% to clearly communicate how each component of the Transformer contributes to generating the probability distribution, and how that distribution---when shaped by sampling hyperparameters---is used to determine the next token.
The key challenge is therefore not only to explain the components that produce the pre-sampling distribution, but to make the entire sampling pipeline observable and learnable as it happens: how logits are formed, how they are transformed by sampling hyperparameters, and how a next token is selected from the resulting distribution.
% 
% Addressing this challenge requires a realtime performant model that can surface dynamically changing outputs under different sampling parameters, along with tightly coupled visualization and interaction, making the problem both conceptually and technically challenging \todo{\cite{}}.
This demands a seamless integration of live inference, probabilistic visualization, and hyperparameter manipulation, which are absent in existing Transformer tutorials or earlier explainer tools. 
% It involves live inference on a real model that reveals the pipeline’s hidden intermediate values, providing immediate and low-latency feedback when learners modify prompts or adjust hyperparameters. Additionally, animations are incorporated to maintain token-level data flow, enabling learners to track changes without losing the overall narrative.

% tool’s tight integration of live inference, probabilistic visualization, and hyperparameter manipulation is not present in existing Transformer tutorials or earlier explainer tools 

%

    \item[C4.] \textbf{Deployment for Scalable Iterative Learning.
    } \label{challenge:access}
    Most educational resources for deep learning tend to rely on static content or provide limited interactivity (e.g., \cite{alammar} for \tf{}). 
    This limitation stems largely from the technical challenge of hosting a live \tf{} model in-browser---these models are large and computationally intensive, making it difficult to achieve the low latency required for real-time interaction \cite{ruan2024webllm}.
    As a result, existing tools tend to rely on pre-selected examples \cite{llm3d, lee2024diffusion} or offer restricted outputs \cite{CNNExplainer}, 
    limiting educational opportunities for interactive hands-on learning, ultimately hindering beginners from gaining deeper understanding and engagement.

\end{enumerate}}
{\begin{enumerate}[label=\textbf{C\arabic*.}, leftmargin=*, ref=\textbf{C\arabic*}]
    \item \textbf{Understanding How Input Text Is Processed Across Complex Model Structures.}\label{challenge:flow}
\tfs{} are complex models composed of many components, such as multi-head self-attention and multi-layer perceptron (MLP), repeated across multiple layers \cite{10.5555/3295222.3295349, ganesh2021compressing} (\autoref{fig:diagram}).
While existing resources have attempted to provide an overview of the model \cite{alammar}, 
they either present all details at once \cite{llm3d}, which may overwhelm beginners, or display model components in disconnected views, often using disjoint visual encodings that increase users’ cognitive load \cite{tufte1983visual}.
However, research shows that presenting  all layer operations and their connections in a unified view has the potential to help users better follow how input data (i.e., token embedding) is transformed into final predictions.
Such attempts typically use diagram-style layouts that place each layer or component as a boxed node and connect them with a line, which works reasonably well for vision models because each node can be visualized as an image \cite{CNNExplainer, kahng2019gan, karpathyConvNetJSMNISTDemo2016, lee2024diffusion}. 
In contrast, Transformers operate on token representations and do not follow the relatively simple feedforward structure of classification models \cite{CNNExplainer, tensorflowplayground, karpathyConvNetJSMNISTDemo2016}. As token representations are interactively transformed as they go through attention, visualizing their transformations introduces unique challenges.
There have been attempts to apply similar diagram-style approaches to Transformers, using component-level boxes connected by simple lines \cite{gao2023transforlearn}, but, they fail to foreground tokens as the primary and coherent visual element and cannot convey an end-to-end transformation path of how each token representation evolves throughout the model.
Hence, an innovative visualization is needed to create a visual summary of the \tf{} that maintains a connected view and preserves data flow.
Adopting an information-flow perspective has potential to help non-experts conceptually link inputs, intermediate computations, and outputs into a coherent process \cite{ferrando2024information}.

\item \textbf{Mathematical Complexity in Multi-Head Self-Attention.}
\label{challenge:math}
Non-experts often struggle to understand the underlying operations in deep learning models \cite{tensorflowplayground}.
In models like CNNs for images, operations can be more easily understood via visual metaphors---such as filters sliding over input images \cite{CNNExplainer}.
In contrast, \tfs{} operate on high-dimensional numerical representations of text, which are less interpretable on their own \cite{NIU202148}.
The complexity deepens with operations like self-attention, where every token interacts with every other token, and with attention-specific operations such as projecting embeddings into Q, K, and V vectors and partitioning them into multiple heads \cite{10.5555/3295222.3295349, Baan2019UnderstandingMA}.
This multi-head attention is foundational to how the model selects, gathers, and combines relevant information from different perspectives \cite{10.5555/3295222.3295349}.
Understanding multi-head attention therefore helps learners understand the concrete mechanism in which words are used to predict the next token, demystifying how the model uses context. 
% 
% What looks like a Transformer “understanding context” actually arises from each token using multiple attention heads to select, gather, and combine relevant information from different perspectives. Understanding multi-head attention therefore helps learners explain which words the model relied on to predict the next token, and why it sometimes misses key context or forms spurious connections, in terms of concrete mechanisms rather than treating the model as a black box.
% 
Existing resources typically explain these operations separately and in detail, but do not effectively help non-experts visually connect how the components work together to transform the token representations \cite{alammar, gao2023transforlearn}. 
Moreover, detailed mathematical explanations are often foregrounded, which can overwhelm non-expert learners with limited mathematical background and discourage further engagement. This gap highlights the need for more accessible explanation, motivating novel visualization and interaction techniques that can unpack and clarify these operations.

 \item \textbf{Understanding Hyperparameters' Impact on Prediction Variability.} 
\label{challenge:parameters}
\tf{} models generate a probability distribution over possible tokens, from which the next token is sampled during autoregressive generation.
Yet many educational resources either fix the prompt and the output, or they omit how the distribution is constructed and how sampling hyperparameters (e.g., temperature, top-$k$, top-$p$) reshape both the candidate set and the final choice \cite{llm3d, gao2023transforlearn}.
As a result, many learners remain unaware of these underlying mechanisms, often viewing \tfs{} as magical or even anthropomorphizing them \cite{stochasticparrots}.
% This highlights a key research challenge: 
% to clearly communicate how each component of the Transformer contributes to generating the probability distribution, and how that distribution---when shaped by sampling hyperparameters---is used to determine the next token.
The key challenge is therefore not only to explain the components that produce the pre-sampling distribution, but to make the entire sampling pipeline observable and learnable as it happens: how logits are formed, how they are transformed by sampling hyperparameters, and how a next token is selected from the resulting distribution.
% 
% Addressing this challenge requires a realtime performant model that can surface dynamically changing outputs under different sampling parameters, along with tightly coupled visualization and interaction, making the problem both conceptually and technically challenging \todo{\cite{}}.
This demands a seamless integration of live inference, probabilistic visualization, and hyperparameter manipulation, which are absent in existing Transformer tutorials or earlier explainer tools. 
% It involves live inference on a real model that reveals the pipeline’s hidden intermediate values, providing immediate and low-latency feedback when learners modify prompts or adjust hyperparameters. Additionally, animations are incorporated to maintain token-level data flow, enabling learners to track changes without losing the overall narrative.

% tool’s tight integration of live inference, probabilistic visualization, and hyperparameter manipulation is not present in existing Transformer tutorials or earlier explainer tools 

%

    \item \textbf{Deployment for Scalable Iterative Learning.
    } \label{challenge:access}
    Most educational resources for deep learning tend to rely on static content or provide limited interactivity (e.g., \cite{alammar} for \tf{}). 
    This limitation stems largely from the technical challenge of hosting a live \tf{} model in-browser---these models are large and computationally intensive, making it difficult to achieve the low latency required for real-time interaction \cite{ruan2024webllm}.
    As a result, existing tools tend to rely on pre-selected examples \cite{llm3d, lee2024diffusion} or offer restricted outputs \cite{CNNExplainer}, 
    limiting educational opportunities for interactive hands-on learning, ultimately hindering beginners from gaining deeper understanding and engagement.

\end{enumerate}}
\section{Design Goals}
\label{sec:design_goals}
Based on the design challenges, we distill four design goals (G1-G4) for \tool{}, an interactive visualization tool to help non-experts learn and experiment with 
\tf{} models.

\aptLtoX{{\begin{enumerate}[topsep=1pt, itemsep=0mm, parsep=1pt, leftmargin=19pt, label=\textbf{G\arabic*.}, ref=\textbf{G\arabic*}]
    \item[G1.] \textbf{Model Overview Prioritizing Token-Centric Data Flow.}\label{goal:flow}
We aim to create a visual summary of the \tf{} architecture as a single, input token centered flow (\ref{challenge:flow}). 
We visualize each token embedding as a one-dimensional heatmap that serves as the primary, consistently used visual element throughout the model. We animate this embedding as it travels along a single continuous path through the architecture, branching and merging in attention, expanding in the MLP, and passing through repeated blocks. Band width is proportional to embedding dimensions, directly revealing how the input evolves.
This innovative design, the first for explaining Transformers, draws inspiration from the Sankey diagram \cite{riehmann2005interactive}, 
which is designed to communicate flow (visually encoded via its edges) between components (via nodes); 
we adapt it to emphasize how information is transformed within the model, building on recent studies that view \tfs{} as dynamic systems \cite{dutta2021redesigning}. 
Our Sankey diagram-inspired design helps users see how input information ``flows'' through the various components and repeated blocks of the \tf{}, undergoing successive processing and transformations before reaching the final output (\autoref{sec:overview}).

    \item[G2.] \label{goal:math}
    \textbf{Visual Disambiguation of Multi-Head Self-Attention with Step-by-Step Visual Explanations.} 
    Given that the multi-head self-attention mechanism is considered the most important yet complex component of the \tf{} \cite{10.5555/3295222.3295349} (\autoref{sec:background}), our goal is to visually unpack how the input embedding is projected into Q, K, and V, branches across heads, participates in attention, and subsequently rejoins through the path within a token-centered animated flow (\ref{challenge:math}), allowing non-experts to form an intuitive visual mental model.
    At the same time, we adopt a progressive disclosure technique \cite{tidwell2010designing} to support learners who want deeper detail. We first present the overall model structure, while leaving detailed mathematical operations to be revealed dynamically through user interaction (\autoref{sec:expandedView}).
    In the detail view, we employ step-by-step animated visualizations to explain the underlying mathematical operations,
    presenting intermediate, successive steps converging toward a final output, inspired by prior research on algorithm visualizations showing that such steps help learners build mental models \cite{Fouh01012012}.
    For example, in the self-attention, we aim to gradually reveal its intermediate steps through animations and visualizations, allowing users to hover over each matrix element to view its value interactively.
    For mathematical operations such as matrix multiplication, we aim to animate how matrices interact, enabling users to trace the computation of each output element (\autoref{sec:expandedView}).

    \item[G3.] \label{goal:experiment} \textbf{Dynamic Experimentation Through User-Provided Text and Hyperparameter Manipulation.}
    To help users understand how the next token is selected from the probability distribution generated by the \tf{} (\ref{challenge:parameters}), we aim to run a live model in browser to support an interactive interface that allows real-time adjustment of sampling hyperparameters (\autoref{sec:sampling}) and to visualize the entire sampling pipeline with real-time intermediate values, so their influence on the next token selection is explicit and understandable.
    The user-provided text input is directly applied in the model's generation process, with their chosen hyperparameters used to predict the next token (\autoref{sec:generation}).
    By experimenting with these settings, users can observe how the output becomes more predictable or more random.
    Predicted tokens are then appended to the input, allowing users to continue generating subsequent tokens and see how early sampling choices propagate through later predictions, while flow-preserving animations maintain token-level data flow across updates.
    This tight integration of live inference, probabilistic visualization, and hyperparameter manipulation helps users understand that the model is not magic, but rather follows a well-defined sequence of operations.

    \item[G4.] \label{goal:web} \textbf{Web-based Tool Powered by Live Model for Interactive Learning.} 
To broaden the public's education access to our tool (\ref{challenge:access}), we aim to build a web-based application that hosts a live, fully-functional \tf{} model. Users would directly interact with the model in their browsers, eliminating the need for installations or specialized hardware. Additionally, to encourage future research and educational use, we open-source our code. (\autoref{sec:implementation})

\end{enumerate}}}
{\begin{enumerate}[topsep=1pt, itemsep=0mm, parsep=1pt, leftmargin=19pt, label=\textbf{G\arabic*.}, ref=\textbf{G\arabic*}]
    \item  \textbf{Model Overview Prioritizing Token-Centric Data Flow.}\label{goal:flow}
We aim to create a visual summary of the \tf{} architecture as a single, input token centered flow (\ref{challenge:flow}). 
We visualize each token embedding as a one-dimensional heatmap that serves as the primary, consistently used visual element throughout the model. We animate this embedding as it travels along a single continuous path through the architecture, branching and merging in attention, expanding in the MLP, and passing through repeated blocks. Band width is proportional to embedding dimensions, directly revealing how the input evolves.
This innovative design, the first for explaining Transformers, draws inspiration from the Sankey diagram \cite{riehmann2005interactive}, 
which is designed to communicate flow (visually encoded via its edges) between components (via nodes); 
we adapt it to emphasize how information is transformed within the model, building on recent studies that view \tfs{} as dynamic systems \cite{dutta2021redesigning}. 
Our Sankey diagram-inspired design helps users see how input information ``flows'' through the various components and repeated blocks of the \tf{}, undergoing successive processing and transformations before reaching the final output (\autoref{sec:overview}).

    \item \label{goal:math}
    \textbf{Visual Disambiguation of Multi-Head Self-Attention with Step-by-Step Visual Explanations.} 
    Given that the multi-head self-attention mechanism is considered the most important yet complex component of the \tf{} \cite{10.5555/3295222.3295349} (\autoref{sec:background}), our goal is to visually unpack how the input embedding is projected into Q, K, and V, branches across heads, participates in attention, and subsequently rejoins through the path within a token-centered animated flow (\ref{challenge:math}), allowing non-experts to form an intuitive visual mental model.
    At the same time, we adopt a progressive disclosure technique \cite{tidwell2010designing} to support learners who want deeper detail. We first present the overall model structure, while leaving detailed mathematical operations to be revealed dynamically through user interaction (\autoref{sec:expandedView}).
    In the detail view, we employ step-by-step animated visualizations to explain the underlying mathematical operations,
    presenting intermediate, successive steps converging toward a final output, inspired by prior research on algorithm visualizations showing that such steps help learners build mental models \cite{Fouh01012012}.
    For example, in the self-attention, we aim to gradually reveal its intermediate steps through animations and visualizations, allowing users to hover over each matrix element to view its value interactively.
    For mathematical operations such as matrix multiplication, we aim to animate how matrices interact, enabling users to trace the computation of each output element (\autoref{sec:expandedView}).

    \item \label{goal:experiment} \textbf{Dynamic Experimentation Through User-Provided Text and Hyperparameter Manipulation.}
    To help users understand how the next token is selected from the probability distribution generated by the \tf{} (\ref{challenge:parameters}), we aim to run a live model in browser to support an interactive interface that allows real-time adjustment of sampling hyperparameters (\autoref{sec:sampling}) and to visualize the entire sampling pipeline with real-time intermediate values, so their influence on the next token selection is explicit and understandable.
    The user-provided text input is directly applied in the model's generation process, with their chosen hyperparameters used to predict the next token (\autoref{sec:generation}).
    By experimenting with these settings, users can observe how the output becomes more predictable or more random.
    Predicted tokens are then appended to the input, allowing users to continue generating subsequent tokens and see how early sampling choices propagate through later predictions, while flow-preserving animations maintain token-level data flow across updates.
    This tight integration of live inference, probabilistic visualization, and hyperparameter manipulation helps users understand that the model is not magic, but rather follows a well-defined sequence of operations.

    \item \label{goal:web} \textbf{Web-based Tool Powered by Live Model for Interactive Learning.} 
To broaden the public's education access to our tool (\ref{challenge:access}), we aim to build a web-based application that hosts a live, fully-functional \tf{} model. Users would directly interact with the model in their browsers, eliminating the need for installations or specialized hardware. Additionally, to encourage future research and educational use, we open-source our code. (\autoref{sec:implementation})

\end{enumerate}}

\begin{figure}[b]
  \includegraphics[width=\linewidth]{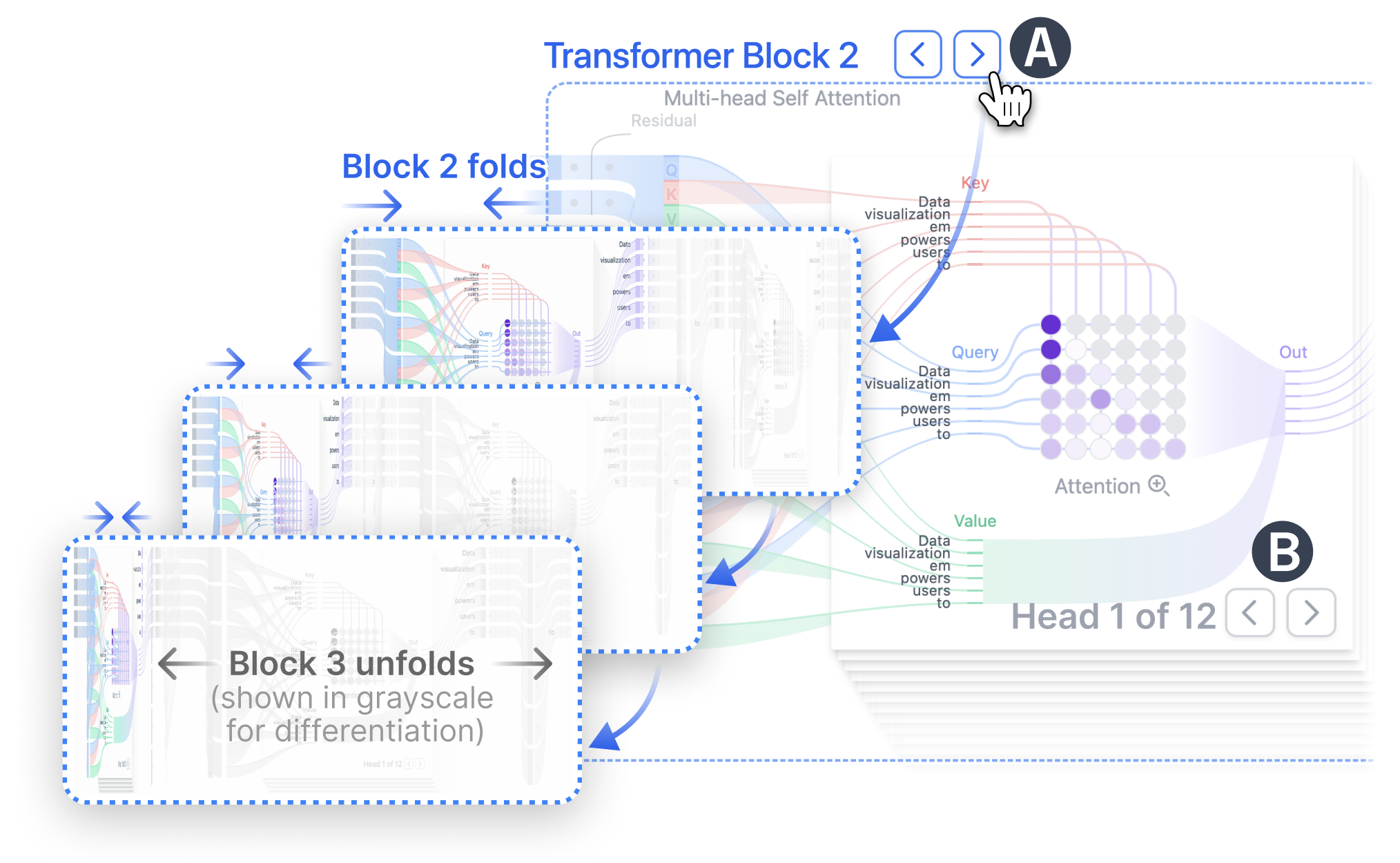}
  \caption{\textbf{(A)} Navigating between \tf{} blocks triggers an animation that folds the current block's flow while unfolding the next. 
  \textbf{(B)} Attention heads are depicted as a stack of cards, with transitions cycling smoothly through the stack in a looping animation. Buttons and annotations enlarged for clarity.}
  \label{fig:pagination}
  \Description{Figure that shows how Transformer Explainer visualizes navigation across blocks and attention heads. Panel A demonstrates that when users move between Transformer blocks, the interface animates the flow by folding the current block while unfolding the next one, illustrated here with Block 2 folding and Block 3 unfolding. Panel B depicts attention heads as a stack of cards, with smooth transitions cycling through each head in a looping animation. Buttons and annotations are shown enlarged for clarity, helping users follow the transitions across blocks and attention heads.}
\end{figure}

\section{Visualization Interface of \tool{}}
\label{sec:interface}
\tool{} visualizes how a trained \tf{} model transforms input text into probabilities for the next-token prediction.
Users can explore the model at different levels of abstraction through \textit{Overview} (\autoref{sec:overview}) and \textit{Step-by-Step Expanded Explanations} (\autoref{sec:expandedView}).
A live GPT-2 \tf{} model running in the user's browser allows real-time experimentation with custom text inputs (\autoref{sec:generation}) and sampling hyperparameters (\autoref{sec:sampling}), enabling users to immediately observe
how these modifications influence the next-token prediction. Our system is targeted towards non-experts, visually guiding them through the mathematical operations underlying a \tf{} model during text generation.
\subsection{Overview}
\label{sec:overview}
\tool{}'s visual design draws inspiration from the Sankey diagram, effective for visualizing data flow~\cite{riehmann2005interactive}, to communicate a high-level overview of how input data flows through the \tf{} model (\ref{goal:flow}; \autoref{fig:teaser}).
Gradient-colored paths illustrate transformations of token embedding vectors across model components, allowing users to understand the model’s structure from a data-flow perspective. 
Vectors are visualized as vertical bars scaled to their actual dimensionality, with a 1D heatmap revealed on hover. These visuals act as illustrative symbols that convey vector shape while achieving a balance between conceptual understanding and visual complexity.
The color mappings are intentionally designed to reflect the role of each component. Token embeddings are rendered in grayscale to emphasize that they are the untransformed initial representations. Q, K, and V use distinct RGB-based colors to clearly differentiate their roles; for example, Q is blue and K is red, and the resulting attention scores appear in purple, visually reflecting the combination of the two. The MLP expands representations, so we preserve a consistent blue gradient to highlight continuity rather than introduce new semantic colors.

The dense \tf{} architecture---consisting of many blocks, each containing multiple attention heads---is visually simplified by displaying only one selected block and one attention head at a time, reducing information overload \cite{llm3d}.
Users can navigate between different blocks and attention heads using pagination buttons \raisebox{-0.2\height}{\includegraphics[height=1em]{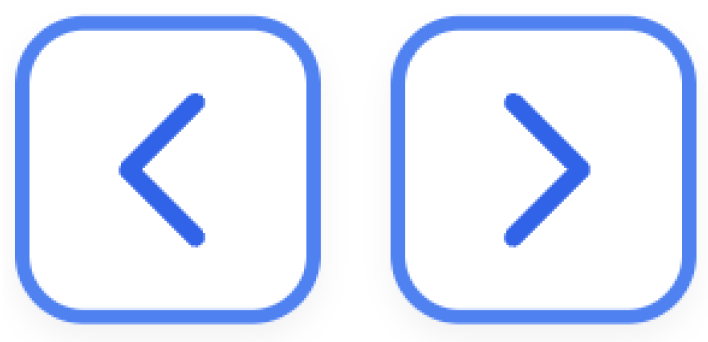}}.
To visually convey the presence of additional blocks and heads beyond the currently displayed one, we use visual metaphors and animated transitions. Repeated \tf{} blocks are represented with gradually fading data flows between them \raisebox{-0.2\height}{\includegraphics[height=1em]{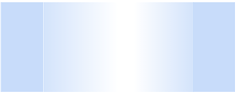}}, subtly suggesting continuity. When navigating between blocks, an animated transition smoothly folds the current block's flow while simultaneously unfolding the next block’s flow into view (\autoref{fig:pagination}A). Attention heads within each block are represented as a stack of cards (\autoref{fig:pagination}B). Transitioning between heads triggers a looping animation, in which cards cycle through the stack. These animated transitions help keep viewers oriented and facilitate the perception of changes \cite{tversky2002animation}.

\subsection{Step-by-Step Expanded Explanations}
\label{sec:expandedView}

We design \textit{Step-by-Step Expanded Explanation Views} to
visualize a model component's internal computations, which %
involve multiple steps with intermediate results. For brevity, we call them \textit{Expanded Views}.
To avoid overwhelming users with details presented all at once (\ref{goal:math}),
our tool enables  users to interactively
open these views by clicking a component that has a magnifying glass icon \raisebox{-0.1\height}{\includegraphics[height=1em]{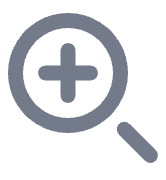}} next to its title. This triggers a smooth animated expansion of the component that preserves the high-level model structure while gradually fading out surrounding areas.
This transition enables the details of the selected component to be presented, while maintaining the high-level context of data flow. %

% \setlength{\columnsep}{6pt}%
% \setlength{\intextsep}{0pt}%
% \begin{wrapfigure}{R}{0.4\textwidth}
%     %
%     \centering
%     \includegraphics[width=0.4\textwidth]{figs/attention-1.5x.png}
% \end{wrapfigure}
\textbf{Self-Attention}.
The \textit{Expanded Self-Attention View} animates the computation of attention scores in three sequential steps to help users understand how each mathematical operation transforms values into final attention scores, and to enable side-by-side comparison of intermediate results, which reduces cognitive load when tracking changes across steps \cite{jang2011spatially}. 
When the user clicks \raisebox{-0.2\height}{\includegraphics[height=1.1em]{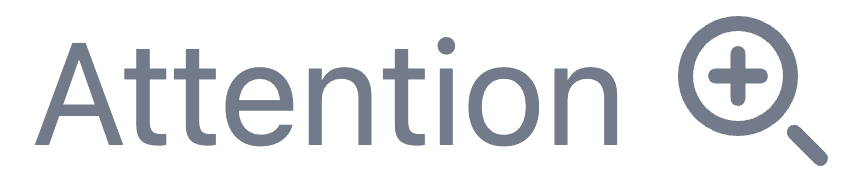}} (\autoref{fig:teaser}:~C2), the flows through the \textbf{\textcolor{keyRed}{key}} and \textbf{\textcolor{queryBlue}{query}} come together to perform the dot product, producing an intermediate matrix. Next, we duplicate this intermediate matrix and present it next to the original one, and show the animation of how it is scaled and masked.
% \begin{figure}[H]
%     \centering
%     \includegraphics[width=0.8\linewidth]{figs/attention-1.5x.png}
%     % \vspace{-5}
% \end{figure}    

\begin{figure}[H]
  \centering
  \includegraphics[width=0.7\linewidth]{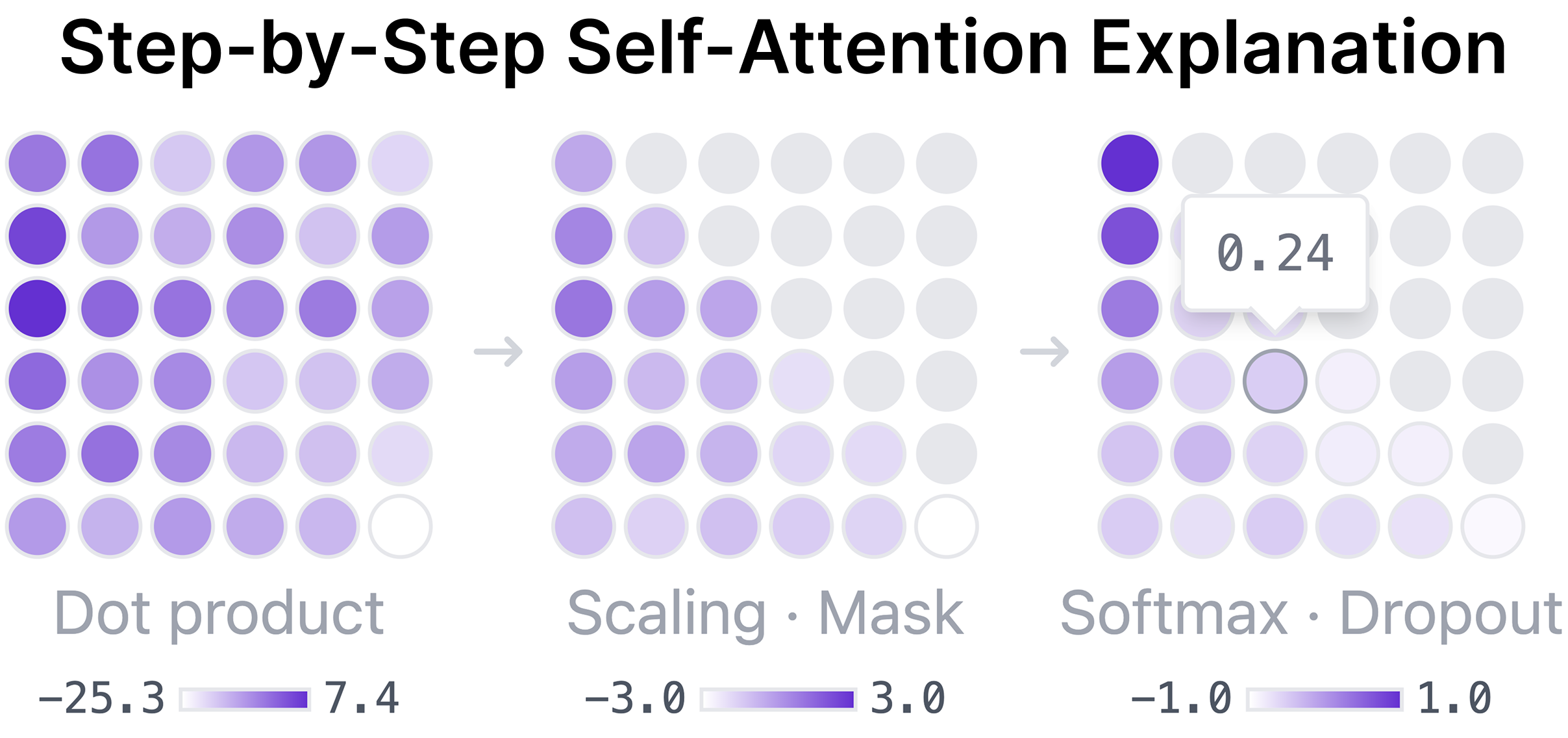}
  \Description{Step-by-step diagram showing how self-attention scores are computed. Three small heatmap matrices illustrate the sequence: dot product, scaling and masking, then softmax and dropout. Each matrix uses a purple color scale to represent numeric values, with darker cells indicating higher attention weights.}
\end{figure}

The same visual process is used to explain %
softmax and dropout, resulting in the final attention scores matrix. %
All matrix values 
are visualized using a heatmap with a purple color scale (e.g., \raisebox{-0.1\height}{\includegraphics[height=0.8em]{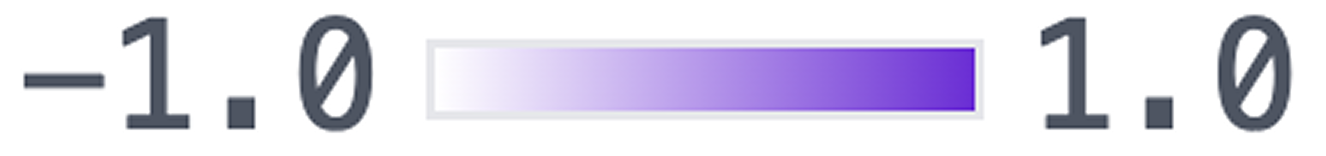}}).
For any matrix output by the three steps, users can hover over individual elements to inspect their numerical values via tooltips.

\begin{figure}[b]
    \centering
    \includegraphics[width=\linewidth]{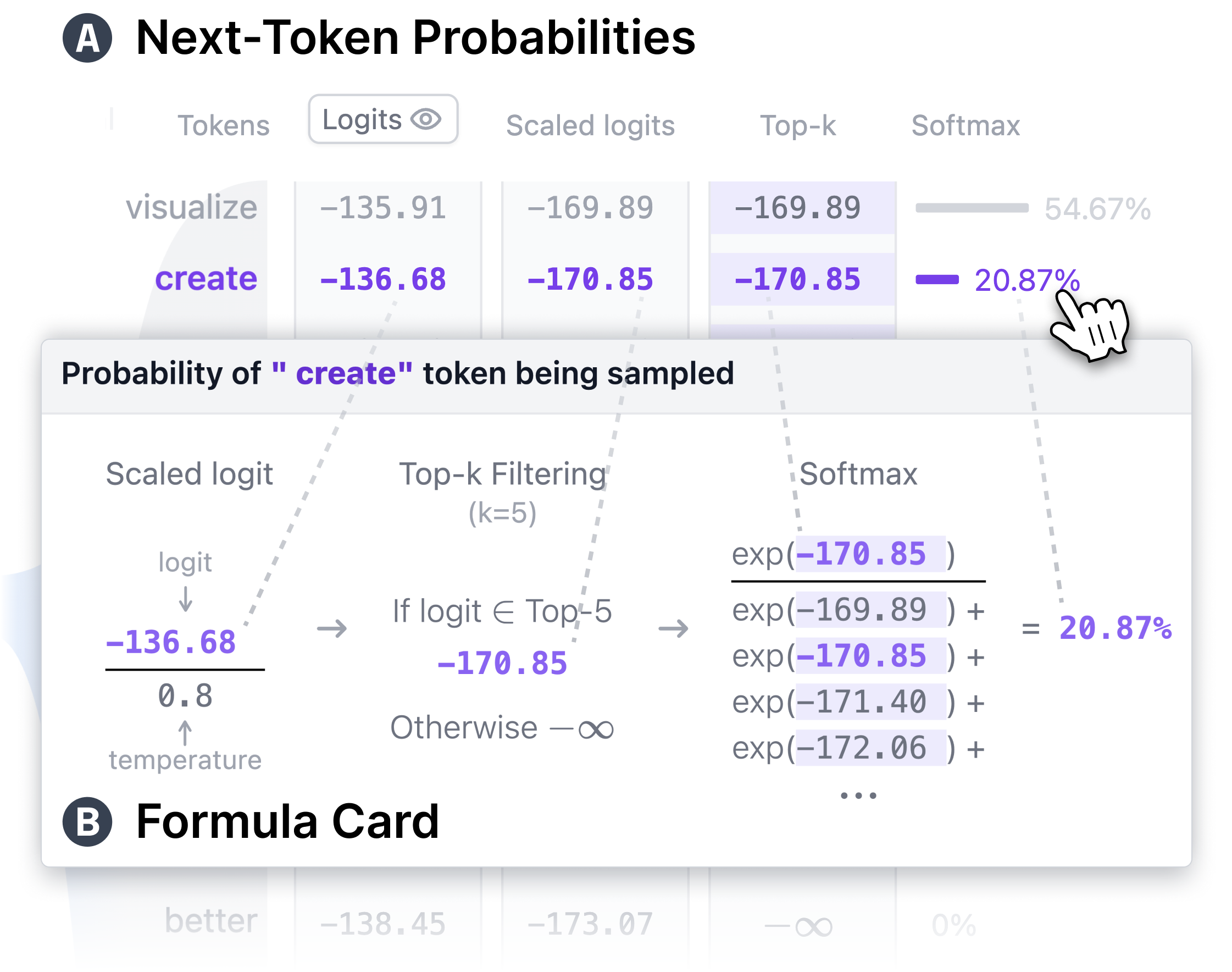}
    \caption{
    \textbf{(A)} A step-by-step expanded explanation view for next-token probabilities, showing the incremental steps in the probability computation. When a user hovers over a token, \textbf{(B)} the formula card updates to show the detailed equations used to compute the probability.
    }
    \Description{Figure that illustrates a step-by-step explanation of how next-token probabilities are calculated in a Transformer. The interface shows tokens with associated logits, scaled logits, and softmax probabilities. In the example, the token “create” is highlighted with a scaled logit of –136.68, and after top-k filtering and softmax computation, it is assigned a probability of 20.87\%. When a user hovers over this token, a formula card appears that expands the underlying calculation, displaying the equations for scaled logits, top-k filtering, and the softmax denominator. This interactive breakdown demonstrates how the system transforms raw logits into normalized probabilities for token selection.}
    \label{fig:probExpanded}
\end{figure}
\begin{figure*}[t]
  \includegraphics[width=1.0\linewidth]{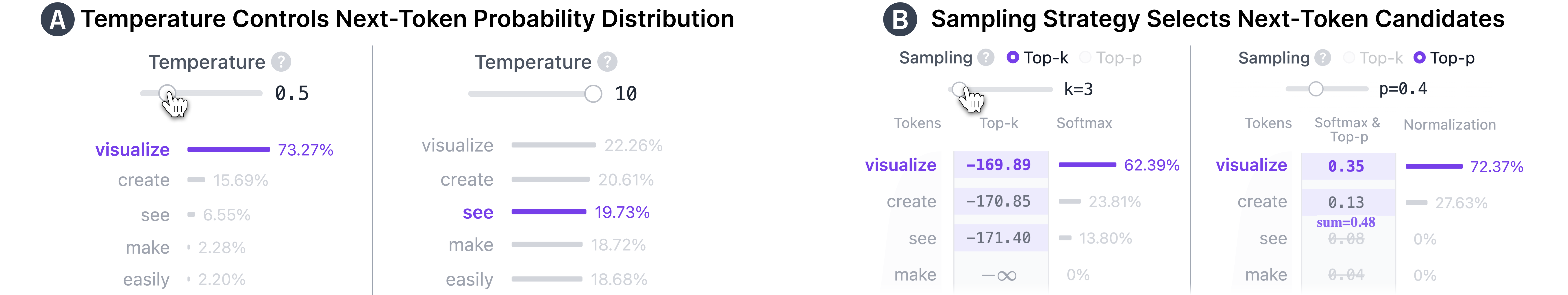}
  \caption{
  \tool{} enables users to adjust inference hyperparameters and observe in real time how they affect next-token prediction. \textbf{(A)} The temperature slider lets users experiment with how temperature shapes the next-token probability distribution. \textbf{A: left}: Lower temperatures sharpen the distribution, making the output more deterministic. \textbf{A: right}: Higher temperatures flatten the distribution, increasing randomness and resulting in less predictable outputs. \textbf{(B)} The sampling strategy selector lets users choose between \textit{top-k} and \textit{top-p}, and adjust the corresponding $k$ or $p$ value using the slider below. \textbf{B: left}: \textit{top-k} sampling with $k = 3$, where the model samples from the top $3$ most likely tokens. \textbf{B: right}: \textit{top-p} sampling with $p = 0.4$, where the model samples from the smallest possible set of tokens whose cumulative probability exceeds $0.4$.
  }
  \label{fig:parameters}
  \Description{Figure that illustrates how Transformer Explainer allows users to manipulate inference hyperparameters such as temperature and sampling strategy to observe their effect on next-token prediction. The left panel (A) shows the impact of temperature on the probability distribution: at a low value of 0.5, the distribution is sharp, making the top token “visualize” dominant at 73.27\%, while at a high value of 10, the probabilities are more evenly spread across tokens such as “visualize,” “create,” and “see,” reflecting greater randomness. The right panel (B) shows the effect of different sampling strategies. With top-k sampling at k = 3, the model chooses among the three most likely tokens, with “visualize” receiving a probability of 62.39\%. With top-p sampling at p = 0.4, the model instead samples from the smallest set of tokens whose cumulative probability exceeds 40\%, assigning 72.37\% to “visualize.” Together, the panels demonstrate how adjusting temperature, top-k, and top-p alters both the shape of the probability distribution and the candidate tokens from which the model selects the next word.}
\end{figure*}

\textbf{Probabilities}.
The \textit{Expanded Probabilities View} (\autoref{fig:probExpanded}A) incrementally displays each step in  probability computation for next-token prediction, from left to right, following how values are transformed as the operation proceeds.
When a user clicks \raisebox{-0.2\height}{\includegraphics[height=1.1em]{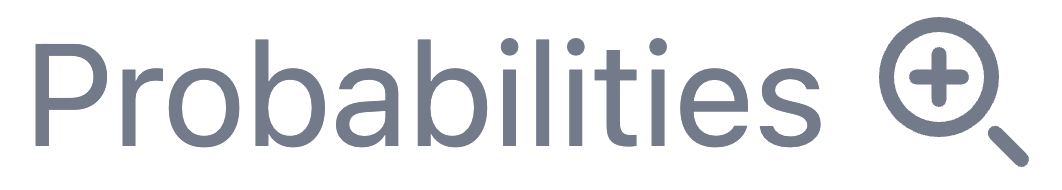}}, which initially shows the final probabilities of next-token candidates (\autoref{fig:teaser}:~C3), the expanded view opens to show each operation and its intermediate values: 
first, the logits (raw prediction scores) are computed from the final embedding;
next, these logits are scaled by the user-selected temperature (\autoref{fig:parameters}A); 
then, the resulting values are filtered based on the user's selected sampling strategy (top-k or top-p) (\autoref{fig:parameters}B), retaining only the most probable candidate tokens---such hyperparameters are important for users to experiment with, as they often cause confusion (\autoref{sec:sampling});
and finally, probabilities for these tokens are computed using softmax.
Tokens are listed by probability, with the most likely candidate at the top.
Hovering over any intermediate value for a token reveals a formula card (\autoref{fig:probExpanded}B) showing the detailed equations used to compute the probability for that specific token at each step.

\begin{figure}[h]
    \centering
    \includegraphics[width=\linewidth]{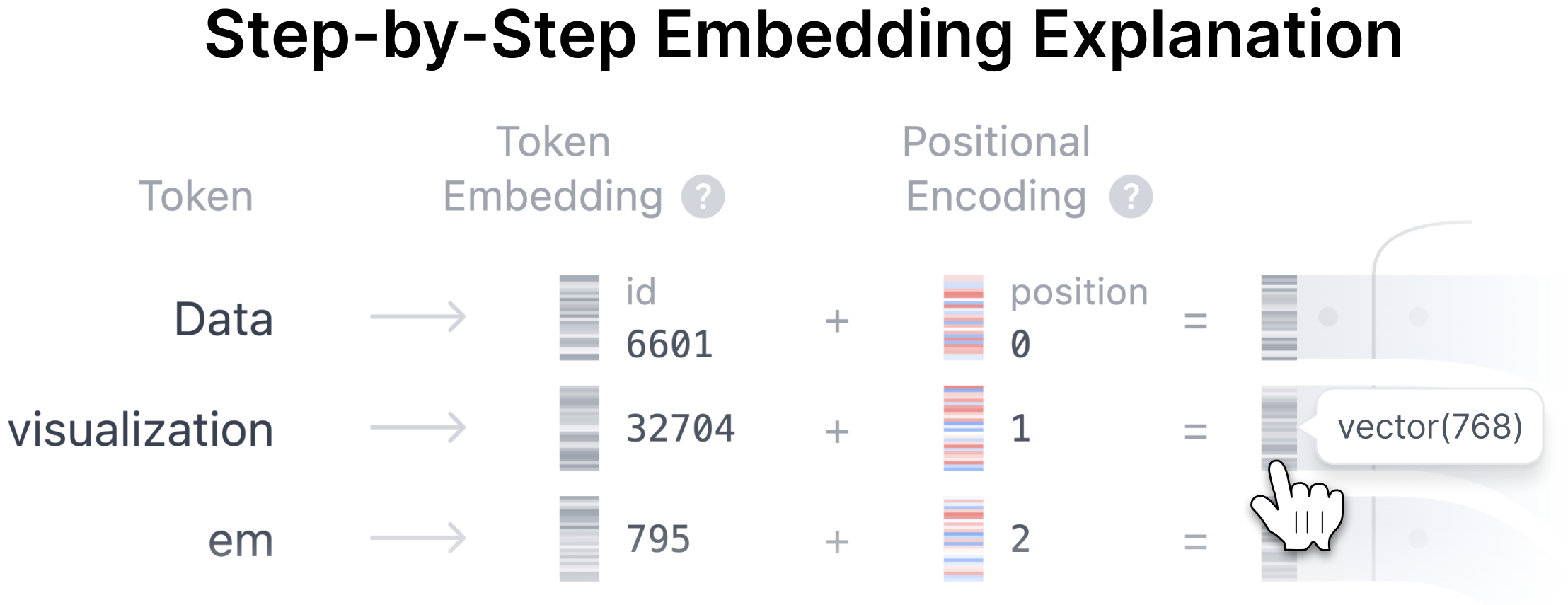}
    \caption{A step-by-step expanded explanation view for embedding visualizes how an input token is converted into its numerical embedding vector.
    }
    \Description{Figure that shows a step-by-step explanation of how tokens are transformed into embedding vectors. The example illustrates three tokens, “Data,” “visualization,” and “em,” each mapped to a unique token ID. These IDs are then converted into token embeddings, represented as high-dimensional numerical vectors. Positional encoding values, such as positions 0, 1, and 2, are added to the embeddings to incorporate sequence order. The resulting outputs are embedding vectors of size 768, visualized here as vertical bars. An interactive cursor highlights the embedding vector, demonstrating how token identity and position information combine to form the numerical representation used by the Transformer.}
    \label{fig:embeddingExpanded}
\end{figure}

\textbf{Embedding}.
The \textit{Expanded Embedding View} (\autoref{fig:embeddingExpanded}) illustrates how each token from the input text is converted into its numerical embedding vector. 
When a user clicks \raisebox{-0.2\height}{\includegraphics[height=1.1em]{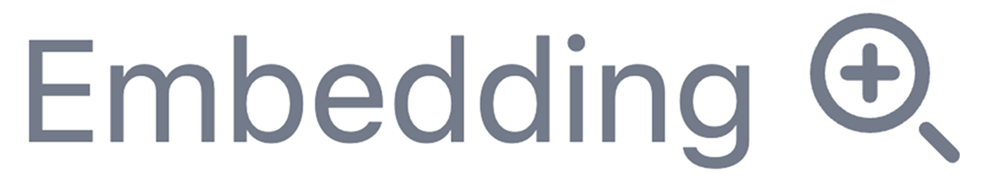}}, the interface displays each token's predefined token ID and position in the sequence, along with its token embedding vector and positional encoding vector, which are summed to produce the final embedding vector that flows into the next component of the model.
Users can inspect all three vectors (i.e., token embedding, positional encoding, summed embedding) as 1-dimensional heatmaps and hover over them to view their dimensions.

\begin{figure}[t]
    \centering
    \includegraphics[width=\linewidth]{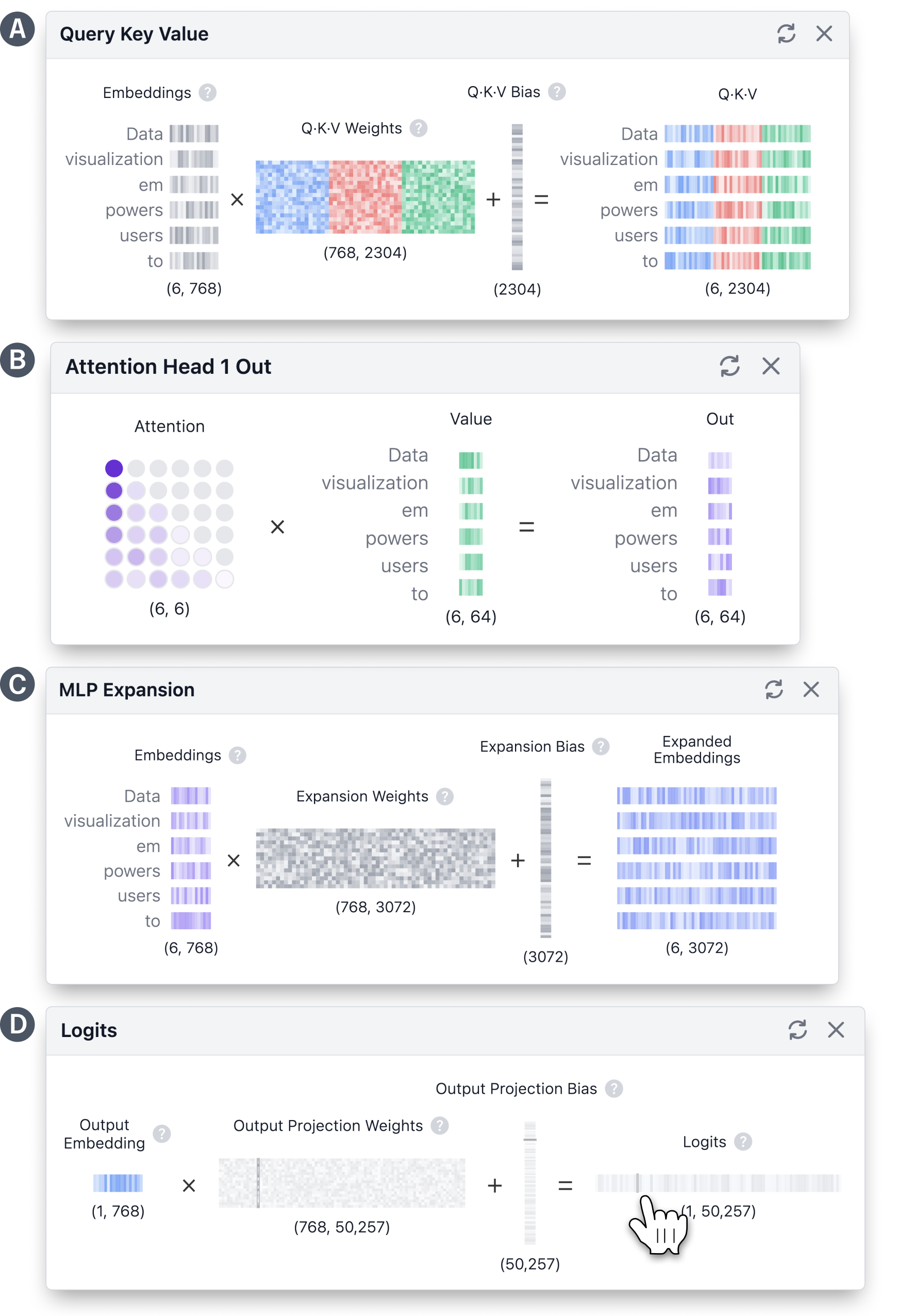}
    \caption{
    \textit{Animated Matrix Multiplication} popovers provide an interactive visualization of embedding-weight multiplication operations within the \tf{} architecture. 
    \textbf{(A)} shows how token embeddings are linearly projected into \textbf{\textcolor{queryBlue}{query}}, \textbf{\textcolor{keyRed}{key}}, and \textbf{\textcolor{valueGreen}{value}} vectors for attention computation. \textbf{(B)} visualizes the multiplication of the attention matrix with value vectors to produce the attention output. \textbf{(C)} shows the dimensional expansion of the embedding through a multi-layer perceptron (MLP). 
    \textbf{(D)} illustrates how the final embedding from the last \tf{} block is multiplied by the output weight matrix to produce logits (raw prediction scores) for next-token prediction. Hovering over an element in the output vector highlights its contributing input elements, helping users understand how specific values are formed.
    }
    \label{fig:math}
    \Description{Figure that presents interactive visualizations of animated matrix multiplication within the Transformer. The first panel shows how token embeddings are projected into query, key, and value vectors by multiplying with a QKV weight matrix and adding bias, resulting in the QKV representation. The second panel illustrates how the attention mechanism multiplies the attention weights with the value vectors to produce the attention output. The third panel depicts the dimensional expansion performed by the multi-layer perceptron, where embeddings are multiplied with expansion weights and bias to create expanded representations. The fourth panel demonstrates how the final embedding is multiplied by the output projection weight matrix and bias to generate logits, which serve as raw prediction scores for the next token. Hovering over any element in the output highlights its contributing input elements, helping users trace how specific values are calculated.}
\end{figure}

\textbf{Animated Matrix Multiplications.} 
Throughout the \tf{} architecture, embeddings repeatedly undergo matrix multiplications with pretrained model weights, changing their dimensions,
reflected in the Sankey diagram paths.
Our tool provides a consistent visualization of these frequently occurring embedding-weight multiplications through popovers (\autoref{fig:math}), helping users recognize that these operations follow the same computational pattern.
Clicking a data flow path between embeddings opens a view that displays the corresponding matrix calculations, animating how input embedding vectors are multiplied by weights to form new embeddings in transformed dimensions. 
Due to the high dimensionality, embeddings and weights are visualized as condensed heatmaps, with exact dimensions displayed below each matrix.
These heatmaps serve as illustrative symbols of vector and matrix computations, maintaining consistent orientations and aspect ratios to convey tensor shape while keeping visual complexity manageable.
Hovering over a specific element in the resulting matrix highlights the contributing inputs (\autoref{fig:math}D), clarifying the calculation details.

\subsection{Real-time Inference for Next-Token Prediction}
\label{sec:generation}
% \setlength{\columnsep}{8pt}%
% \setlength{\intextsep}{0pt}%
% \begin{wrapfigure}{r}{0.3\textwidth}
%     \centering
%     \includegraphics[width=0.3\textwidth]{figs/input-2x.png}
% \end{wrapfigure}
Users can enter custom text into the input bar and click \raisebox{-0.2\height}{\includegraphics[height=1.2em]{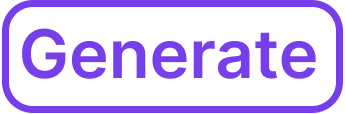}} 
to observe in real-time how the next token is predicted.
The input text is immediately broken into tokens and updated in the visualization;
then, a smooth animation visualizes the updated data flowing through the model. The animation provides continuity between embeddings, clearly showing changes in position, size, shape, and color, helping users track data updates \cite{heer2007animated}.
While this animation plays, a loading indicator appears in the input bar, and the predicted next token is appended once the data flow reaches the output. 
\begin{figure}[H]
    \centering
    \includegraphics[width=\linewidth]{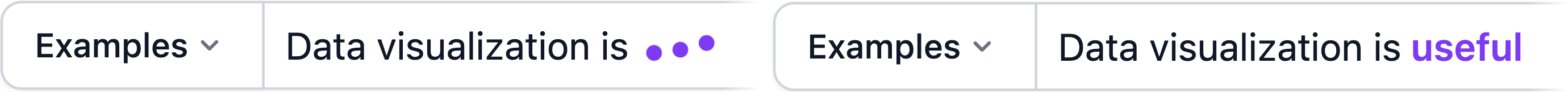}
    \Description{Interface snippet showing an example text-generation bar. A dropdown labeled “Examples” appears next to an input sentence that updates from a partial phrase to a completed sentence, with highlighted tokens indicating generated output.}
\end{figure}
Users can repeatedly click the Generate button to continue generating tokens, visually understanding how a \tf{} model builds sentences one token at a time (\ref{goal:experiment}). 
To support installation-free access, when the user visits our tool, the GPT-2 model (including its weights) is downloaded to the user's browser and runs entirely in it  (\autoref{sec:implementation}). 
To maintain usability in high-latency or low-bandwidth environments where model loading may take time, five example prompts with pre-computed intermediate data extracted from models are provided, allowing users to explore the interface instantly.

\subsection{Adjustable Sampling Hyperparameters Influencing Next-Token Prediction}
\label{sec:sampling}
\tool{} enables users to adjust inference hyperparameters and observe in real-time how these settings influence next-token prediction (\ref{goal:experiment}).
Users can modify temperature, sampling strategies, and their associated hyperparameters using the interactive controls located next to the Generate button (\autoref{fig:teaser}B). By testing their hypotheses and observing immediate feedback, users can see that \tfs{} select the next token based on probabilistic algorithms---not randomly or through ``magic.''

\textbf{Temperature} (\autoref{fig:parameters}A).
The temperature hyperparameter shapes the generated probability distribution for the next-token prediction, making it sharper (lower temperature) or smoother (higher temperature).
Users can adjust the temperature using a slider and test how it affects prediction determinism, understanding that temperature determines whether the output becomes more deterministic or random.

\textbf{Sampling Strategies} (\autoref{fig:parameters}B).
We provide two widely-used sampling strategies: \textit{top-k} and \textit{top-p}.
Users can select among sampling strategies using radio buttons and adjust the corresponding hyperparameters through sliders, observing how these hyperparameters influence which tokens are considered for the next prediction and the likelihood of each token being selected.
Adjustments are reflected instantly, with the \textit{Expanded Probabilities View} displaying how probabilities are computed based on the selected inference hyperparameters.

%

%
%

%
%

%
%
%
%
%

% \setlength{\columnsep}{8pt}%
% \setlength{\intextsep}{0pt}%
% \begin{wrapfigure}{R}{0.2\textwidth}
%     \centering
%     \includegraphics[width=0.2\textwidth]{figs/residual-2x.png}
% \end{wrapfigure}
\subsection{Auxiliary Architectural Features} 
% While advanced architectural features (e.g., layer normalization) enhance \tf{} performance, in-depth understanding of these features is not necessary for grasping the model's core logic and data flow.
% \begin{figure}[H]
%     \centering
%     \includegraphics[width=0.5\linewidth]{figs/residual.png}
% \end{figure}
We treat layer normalization, residual connections, activation functions, and dropout as supporting or conditioning mechanisms: they modulate and stabilize the primary computations (attention mixing and MLP transformations) and help preserve signal across depth, but are not the central conceptual steps we target for explaining how next-token probabilities are produced and sampled. This categorization was informed by consultations with machine learning instructors (\autoref{sec:iteration}), who noted that these mechanisms introduce additional mathematical detail that can overwhelm beginners without an ML background.
To balance complexity, our tool visualizes these auxiliary features using visual scent \cite{willett2007scented}: dots represent layer normalization, dropout, and activations while lines indicate residual connections. 
\begin{figure}[H]
  \centering
  \includegraphics[width=0.7\linewidth]{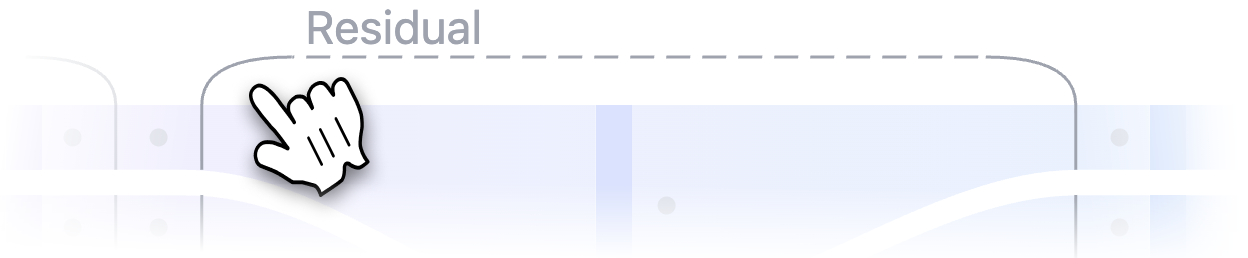}
  \Description{A dashed flowing line labeled “Residual” illustrates a residual connection path across layers.}
\end{figure}
On hovering, residual connections are represented as dashed flowing lines, with animations illustrating the flow direction from their origin to their destination.
Hovering over a symbol displays a brief explanatory tooltip, and users can click the \textit{Read More} button to access a supplementary article located below the tool.

\label{sec:navigation}
\begin{figure}[!t]
  \includegraphics[width=\linewidth]{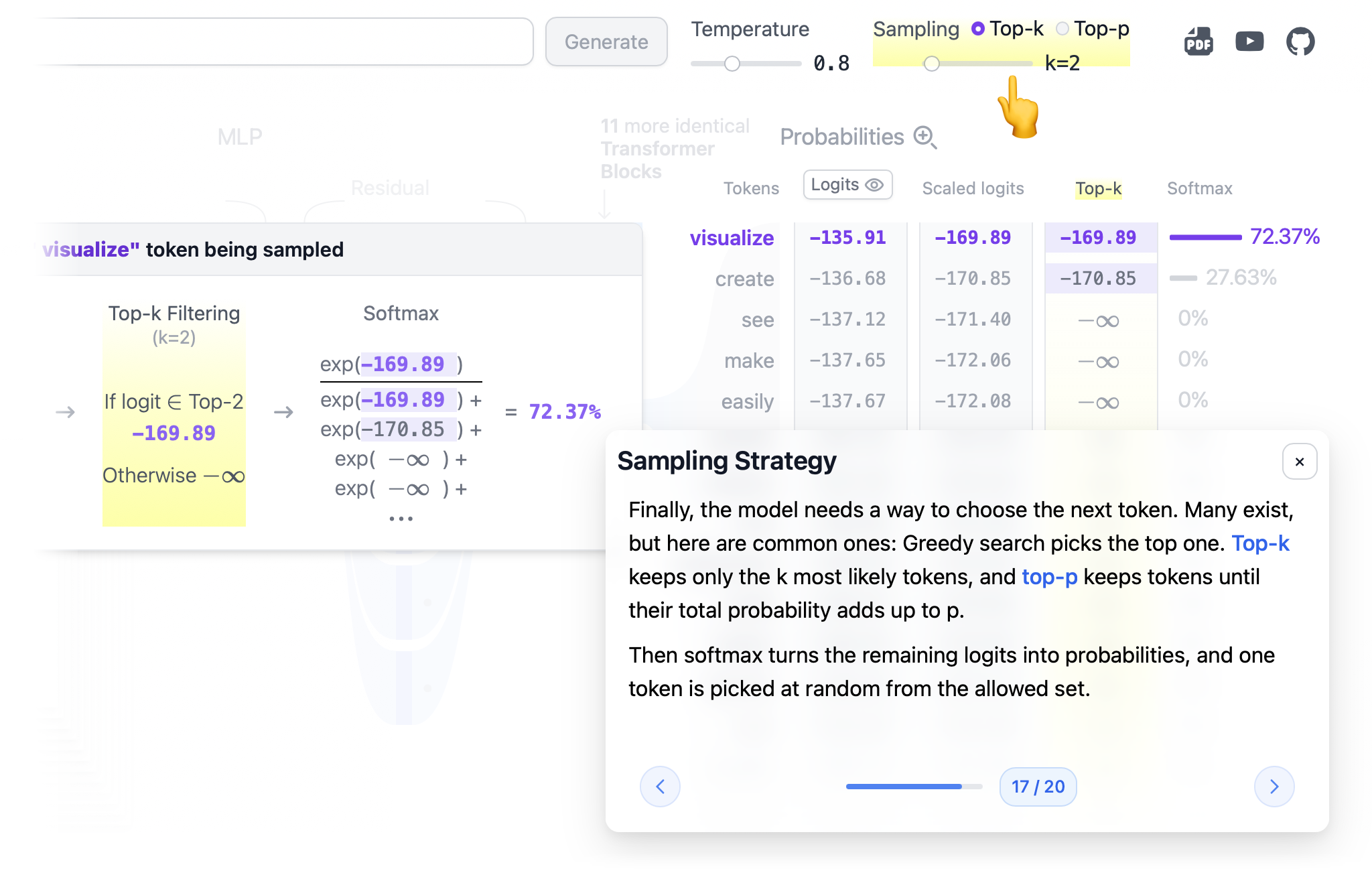}
  \caption{
Guided learning provides an interactive, step-by-step text explanation that introduces Transformer concepts and connects relevant visual components. For example, on the \textit{sampling strategy} page, it guides users to manipulate hyperparameters while simultaneously highlighting the elements that change in response on the screen.
  }
  \label{fig:guided-learning}
  \Description{Figure that shows an example of the guided learning interface explaining sampling strategies in Transformer text generation. The screen highlights the “Top-k” sampling option with k = 2, showing how logits are filtered so that only the top two remain before applying the softmax function. The left panel illustrates the mathematical steps of top-k filtering and probability calculation, with the token “visualize” being sampled at a probability of 72.37\%. On the right, a guided learning text card titled “Sampling Strategy” explains in plain language how greedy search, top-k, and top-p sampling work. The interface visually connects this explanation with corresponding changes in the probability table above, where the highlighted elements update as the user manipulates the hyperparameters. This example demonstrates how guided learning provides a step-by-step link between mathematical formulas, probability values, and explanatory text.}
\end{figure}
\subsection{Guided Learning}
\label{sec:guided-learning}
Guided learning is an interactive text card (\autoref{fig:guided-learning}) that introduces Transformer concepts by following the flow of data, starting from the principle of autoregression and progressively covering the overall architecture and its main components—embedding, attention, MLP, and output probability.

At each step, users can follow the yellow finger icon to click dynamic elements to expand the view into detailed mathematical operations (\autoref{sec:expandedView}), manipulate input text or hyperparameters (\autoref{sec:generation}, \autoref{sec:sampling}), and navigate across blocks and heads (\autoref{sec:overview}). 
The visual elements directly related to the current guided learning page are highlighted in yellow, helping users focus their attention and connect concepts. 
Through this process, users not only learn the core concepts of the Transformer progressively and contextually but also naturally become familiar with the tool’s interactive features.

Guided learning can be accessed at any time via the floating button located at the bottom right of the screen \raisebox{-0.2\height}{\includegraphics[height=1.2em]{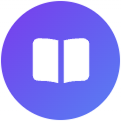}}, and users can move directly to any page using the navigation buttons \raisebox{-0.3\height}{\includegraphics[height=1.2em]{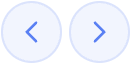}} or the page dropdown \raisebox{-0.3\height}{\includegraphics[height=1.2em]{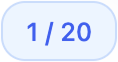}} at the bottom of each card. 
In addition, guided learning also functions as a form of in-situ text explanation: when users hover over visual elements related to Transformer concepts, a help cursor appears, and clicking it opens the corresponding guided learning page. 
This design prevents the context switching  where users would otherwise need to scroll down to the article at the bottom to view an explanation while freely exploring the tool.

The guided learning feature was introduced after a preliminary usability assessment conducted in the second phase of the tool’s design iteration (\autoref{sec:phase2}), reflecting participants’ feedback that an onboarding tutorial and in-situ text explanations could help lower the initial learning curve.

\subsection{Web-Based, Open-Source Implementation}
\label{sec:implementation}
\tool{} is a web-based, open-source visualization tool designed to help non-experts understand how \tfs{} work (\ref{goal:web}).
Users can access our tool using only a web browser, with no installation or specialized hardware required.
We use a HuggingFace \tfs{}' \cite{wolf2020transformers} GPT-2 Small model from NanoGPT \cite{karpathy_nanoGPT} to extract model data used in calculations during inference.
To run the model in the browser, we converted a PyTorch model into ONNX format and used the ONNX Runtime Web API \cite{onnxruntime}.
To minimize loading time, we split the model files into smaller chunks for parallel downloads and cache the model data in the browser using IndexedDB.
As a result, the model only needs to be downloaded once, upon the user's first visit.
The frontend is built with Svelte \cite{svelte} and D3.js \cite{bostock2011d3}.
\section{Informed Design Through Iterations}
\label{sec:iteration}
The current design of our tool is the result of over a year of iterative investigation and development, shaped by feedback gathered across three major phases.

\begin{itemize}
    \item \textbf{Phase 1: Initial Prototype Feedback.} We built an early prototype and collected in-the-wild usage signals, complemented by informal feedback from instructors who regularly teach Transformer-related topics.
    \item \textbf{Phase 2: Enhanced Model Exploration.} 
    Building on the initial feedback, we expanded the tool to support deeper exploration of model internals, enabling users to flexibly navigate the architecture (e.g., across Transformer blocks and attention heads (\autoref{sec:overview}); and experiment with the output generation process through adjustable sampling hyperparameters (\autoref{sec:sampling}). 
    We then conducted a preliminary usability assessment to gather early feedback on the tool's effectiveness as a learning resource.
    \item \textbf{Phase 3: Guided Learning Support.} Drawing on the preliminary feedback, we refined the design into its final form by adding guided learning scaffolds (\autoref{sec:guided-learning}). 
    We then validated the resulting tool through a summative evaluation (\autoref{sec:user_study}).
\end{itemize}

\subsection{Phase 1: Initial Prototype Feedback}
\label{sec:phase1}
The first release of \tool{} implemented our initial design goals (\autoref{sec:design_goals}): a flow–based visualization (\ref{goal:flow}), step-by-step visual explanation (\ref{goal:math}), and dynamic experimentation through user input and hyperparameter manipulation (\ref{goal:experiment}). To manage early complexity, it rendered only the first block and first head—visually signaling repetition but not enabling traversal across blocks or heads.
Following the release, we collected informal feedback from early users and three 
instructors who have taught graduate and undergraduate courses in machine learning and NLP topics (two instructors are co-authors). 
They noted that the multi-head and multi-block concepts remained abstract without the ability to systematically explore individual blocks and heads, and compare attention patterns; they also requested a more granular account of the output probabilities generation process that produces next-token distributions. These observations motivated the next iteration, emphasizing richer navigation across blocks and heads and an explicit, step-wise visualization of the final output probabilities layer.

% \setlength{\columnsep}{8pt}%
% \setlength{\intextsep}{0pt}%
% \begin{wrapfigure}{r}{0.3\textwidth}
%     % \vspace{0pt}
%     \centering
%     \includegraphics[width=0.3\textwidth]{figs/formative-study.png}
%     \vspace{2pt}
% \end{wrapfigure}
\subsection{Phase 2: Enhanced Model Exploration}
\label{sec:phase2}
We extended the tool to support navigation across attention heads and Transformer blocks (\autoref{sec:overview}), and introduced a step-by-step output probabilities formula card (\autoref{sec:expandedView}), including controls for sampling hyperparameters (\autoref{sec:sampling}). 
With these features in place, our tool was ready for a preliminary usability assessment to collect feedback on how its interactivity would benefit non-expert learners.
We used a within-subjects design to compare against a blog post baseline, which was static and non-interactive.  
% \begin{figure}[H]
%     \vspace{1em}
%     \centering
%     \includegraphics[width=0.7\linewidth]{figs/formative-study.png}
%     % \caption{Caption}
%     % \label{fig:placeholder}
%     \vspace{1em}
% \end{figure}

\begin{figure}[H]
  \centering
  \includegraphics[width=0.7\linewidth]{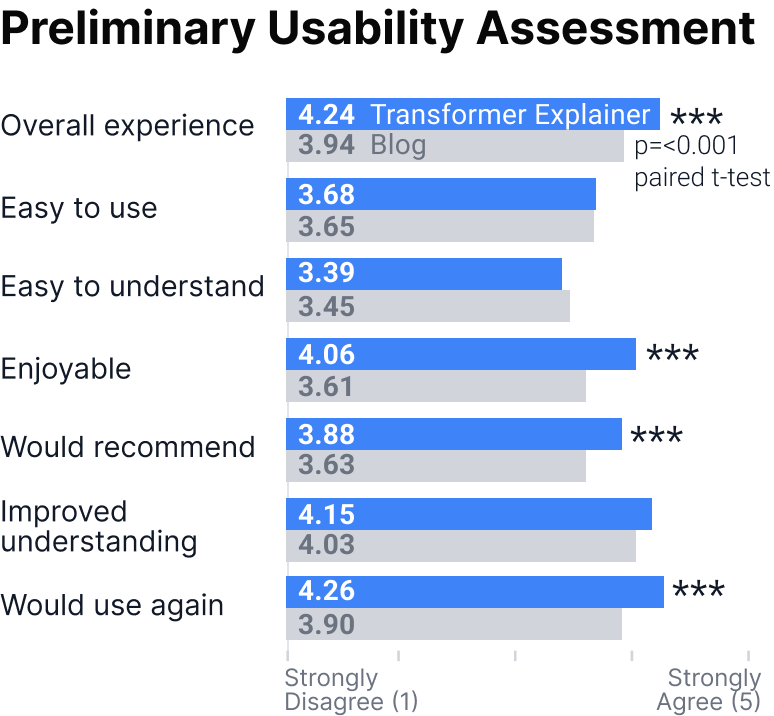}
  \Description{Bar chart comparing usability ratings between Transformer Explainer and a static blog baseline across several measures. Blue bars (Transformer Explainer) are generally higher than gray bars (blog) for experience, enjoyment, recommendation, and reuse, with statistical significance markers shown.}
\end{figure}

A large majority of the 145 study participants (74\%) 
preferred \tool{} over the blog, 
and rated our tool significantly higher on most usability items (as shown in the right figure), including
\textit{overall experience}, \textit{enjoyment}, and \textit{likelihood to recommend} or \textit{use again}. 
These findings provided early evidence that interactive visual exploration and experimentation can better support non-expert learners than static materials.

\subsection{Phase 3: Guided Learning Support}
Through the preliminary usability assessment, we also identified opportunities to improve the tool. 
Transformer Explainer offers rich interactivity and dynamic visualizations; however, these same features sometimes created usability challenges. 
Some participants noted that the initial learning curve could be eased with a more guided onboarding experience. These observations align with usability survey responses, which indicated room for improvement on \textit{“easy to use”} and \textit{“easy to understand.”}

Therefore, we introduced the guided learning (\autoref{fig:guided-learning}, \autoref{sec:guided-learning}), an interactive, step-by-step text card that explains Transformer concepts while linking them to visual components and interactive actions. 
With these updates in place, we conducted a summative between-subjects evaluation to assess the tool’s usability and usefulness, and to examine whether \tool{} improves non-expert learners’ understanding of Transformer concepts compared to blogs and videos, which are popular means of learning. %
\section{Evaluation: User Study}\label{sec:user_study}
We conducted a user study to evaluate how effectively \tool{} meets the design goals identified in \autoref{sec:design_goals} and supports our high-level research contributions (\autoref{sec:intro}).
Specifically, we address three research questions (RQ1-3):

\aptLtoX{\begin{enumerate}[label=\textbf{RQ\arabic*.}, leftmargin=*, ref=\textbf{RQ\arabic*}]
\item[RQ1.] \label{rq:understanding} How does \tool{} improve non-expert learners’ understanding of Transformer concepts compared to blog posts and videos, which are popular means of learning?
\item[RQ2.] \label{rq:experience} 
How is learners' personal experience enhanced when learning Transformer concepts?
\item[RQ3.] \label{rq:feature} How do the features of \tool{} support effective learning?

\end{enumerate}}
{\begin{enumerate}[label=\textbf{RQ\arabic*.}, leftmargin=*, ref=\textbf{RQ\arabic*}]
\item\label{rq:understanding} How does \tool{} improve non-expert learners’ understanding of Transformer concepts compared to blog posts and videos, which are popular means of learning?
\item\label{rq:experience} 
How is learners' personal experience enhanced when learning Transformer concepts?
\item\label{rq:feature} How do the features of \tool{} support effective learning?

\end{enumerate}}

In addition, we examined differences in usability, engagement, and perceived learning experience across educational resources. 

\begin{table*}[t]
\centering
\begin{tabular}{p{0.4\textwidth} p{0.45\textwidth}}
\toprule
\textbf{Learning Objective (LO)} & \textbf{Quiz Question} \\
\midrule
LO1: How GPT-2 generates text one token at a time 
& Q1: How does a Transformer generate text? \\
\addlinespace
LO2: The overall Transformer architecture 
& Q2: Which structure matches a text-generation Transformer? \\
\addlinespace
LO3: Text to Embedding transformation 
& Q3: Which is the best description of tokens and embeddings? \\
\addlinespace
LO4: Multi-head Self-Attention mechanism 
& Q4-1: In self-attention, what do Query, Key, and Value do? \newline
Q4-2: Why use multiple heads in attention?\\
\addlinespace
LO5: MLP (feed-forward network) 
& Q5: Why add a MLP (feed-forward) layer after attention? \\
\addlinespace
LO6: Final probabilities and sampling parameters 
& Q6: How do “top-k” or “temperature” affect text generation? \\
\bottomrule
\end{tabular}
\caption{Learning objectives (LO) and their corresponding multiple-choice quiz questions.}
\label{tab:lo_quiz}
\end{table*}

\subsection{Study Design}
\label{sec:study-design}
We conducted a controlled between-subject experiment to evaluate the learning effectiveness of \tool{} in comparison to existing educational resources. Participants were randomly assigned to one of three conditions: \tool, a blog post, or an educational video. 
A between-subjects design was chosen to minimize knowledge transfer effects across conditions that participants would otherwise experience if they were to go through the conditions in sequence.
This study was approved by our institution’s IRB.

The blog and video baselines were selected because they represent two popular modes of learning resources \cite{ten2015like, navarrete2025closer, dennis2015blogging, shaohui2008application}. 
Blogs are static narrative media, composed of text and figures that learners interpret at their own pace; 
whereas videos are multimodal media that integrate narration, animations, and other elements in a fixed sequence. 
Comparing our tool against these two formats allowed us to examine the added contribution of interactivity and dynamic visualization beyond static or multimodal presentation.
For a fair comparison, we ensured that all three learning modes convey the same information.
Specifically, we selected the blog post on decoder-only \tf{} (GPT-2),\footnote{\url{https://jalammar.github.io/illustrated-gpt2/}} which is part of the blog returned as the top-ranked Google search result.\footnote{For search terms such as ``Transformer explained'' and variations} 
Similarly, we selected the Transformers video series by 3Blue1Brown\footnote{\url{https://youtu.be/wjZofJX0v4M}} which has been viewed over 7.5 million times. 
As the blog post and video contained content irrelevant to Transformer learning in our context (e.g., image generation model rather than text-based model), we consulted with the three instructors (\autoref{sec:phase1}) to identify six key learning objectives (\autoref{tab:lo_quiz}) important for resources helping non-experts gain a conceptual understanding of Transformers. 
After removing irrelevant content, the learning resources could be fully explored in about 30 minutes by a learner.

\subsection{Participants} %
\label{sec:participants}
We recruited participants from Prolific\footnote{\url{https://www.prolific.com}}, an online user study platform, for a 1-hour study. 
The average completion time for the study was 43 minutes, and each participant received compensation of \$12.
To ensure the study targeted non-expert learners, we asked prospective participants to self-rate their familiarity with generative AI on a 5-point scale and to indicate whether they were interested in learning how text-based generative AI works. 
Only individuals who both expressed interest 
and reported low familiarity levels were eligible to participate (i.e., either 1: \textit{Don’t know what it is}; 2: \textit{Heard of it only}; or 3: \textit{Aware but don't understand}). 
90 of them completed the study, balanced across the three experimental conditions. (\autoref{screening} describes the approach for checking participant engagement.)

Overall, the participants spanned a wide range of educational backgrounds, disciplines and industries.
In detail, their educational backgrounds included
bachelor's degree (35.6\%),
master’s degree (18.9\%),
high school (16.7\%),
and some college (13.3\%).
Participants also spanned diverse disciplines, including computer science and information technology (30.0\%)
and business/management (23.3\%), as well as social sciences, arts and humanities, and health sciences. 
Industry backgrounds were varied, with notable representation from health care (14.4\%), information services (13.3\%), technical services (7.8\%), and finance (7.8\%).

\subsection{Procedure}
\label{sec:procedure}

After providing informed consent, participants first completed a \textbf{demographic and background survey}, which included self-ratings of mathematical proficiency and generative AI familiarity. They were then randomly assigned to one of three experimental conditions: interactive tool, blog, or video (referred to as \textcolor{TEColorscale}{\tool{}}, \textcolor{BPColorscale}{Blog}, and \textcolor{VDColorscale}{Video}, respectively).
Participants studied their assigned resource freely for up to 45 minutes,
enough time to fully explore the material (\autoref{sec:study-design}) 
and pace their learning  while keeping  exposure timing comparable across conditions. 
Next, participants completed the post-study evaluation, which included both objective and subjective measures.

\begin{itemize}
    
\item 
For objective assessment, participants took a \textbf{closed-book multiple-choice quiz} with 7 questions (\autoref{tab:lo_quiz}) closely aligned with the learning objectives identified by instructors (\autoref{sec:phase1}), who also helped review the questions. 
The quiz questions focused only on concepts shared across all three resources, and the quiz items and answer choices were identical for all conditions.

\item 
For subjective assessment,
participants (1) reported their \textbf{self-perceived understanding} of Transformer concepts on a 5-point Likert scale and \textbf{rated the learning experience} of their assigned material;
and (2) answered \textbf{three open-ended reflection questions} (``most helpful aspects,'' ``most confusing aspects,'' and ``suggested improvements'') to capture qualitative feedback.
Participants in the \textcolor{TEColorscale}{\tool{}} condition also provided tool-specific \textbf{evaluations of individual features}.

\end{itemize}

\subsubsection{Checking for participant engagement}
\label{screening}
To safeguard the integrity of the study results, we employed a two-pronged verification to ensure participants had used the tools and engaged both meaningfully and honestly.
First, participants were required to answer three simple resource-specific questions that anyone who had studied the resource could easily answer (e.g., ``What is the temperature range that can be adjusted with the slider in the tool?''). Only those with at least two correct responses were included, serving as both an engagement check and a verification of minimal attention. 
Second, we excluded participants showing signs of inattention or dishonesty, such as unrealistically short quiz times ($\leq 7.1$ seconds per question, below the 5th percentile) or insufficient effective tool-use time ($\leq 5$ minutes of focused engagement, discounting tab-outs or early exits). 
Using this two-pronged process, we excluded 38 of 128 initial participants, resulting in a final sample of 90 motivated non-experts.

\begin{figure*}[!t]
  \centering
  \includegraphics[width=1\linewidth]{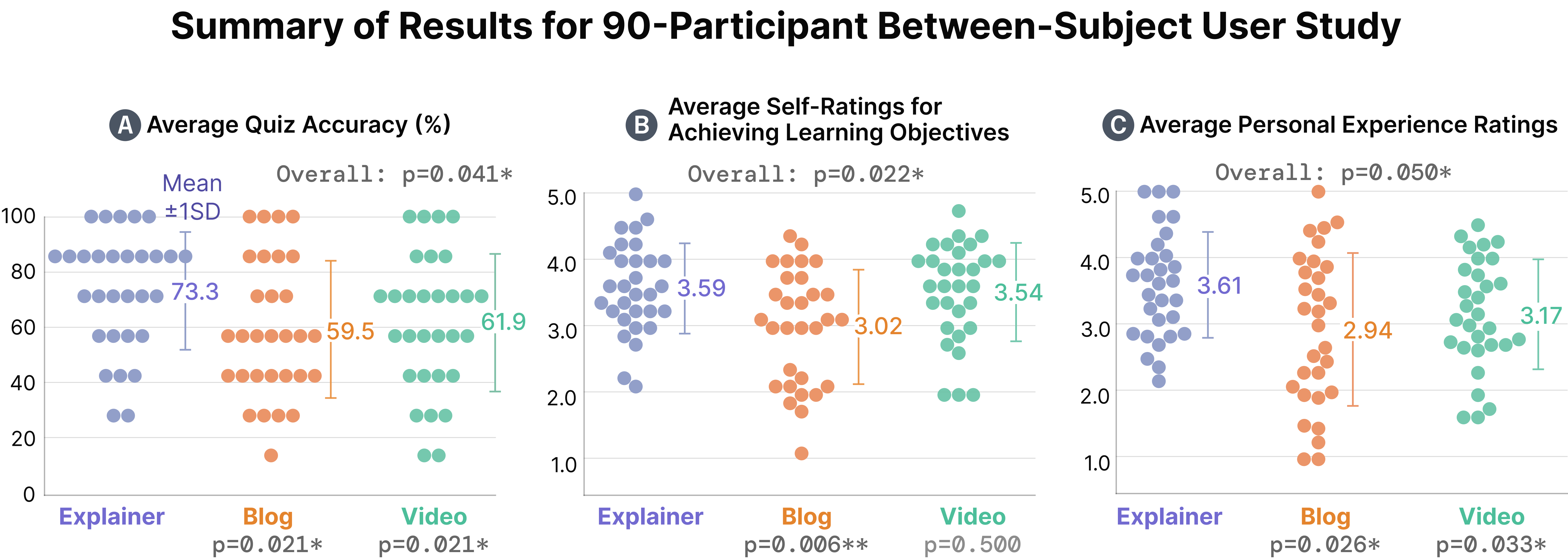}
  \caption{
\textcolor{TEColorscale}{\tool{}} was significantly more effective than both \textcolor{BPColorscale}{Blog} and \textcolor{VDColorscale}{Video} in understanding Transformer concepts and learning experience.
\textbf{(A)} \textcolor{TEColorscale}{\tool{}} participants achieved higher quiz accuracy than \textcolor{BPColorscale}{Blog} ($p=0.021$) and \textcolor{VDColorscale}{Video} ($p=0.021$), as confirmed by pairwise tests from a GLMM controlling for participants’ prior math and AI knowledge.
\textbf{(B)} They also reported higher achievement of learning objectives than \textcolor{BPColorscale}{Blog} ($p=0.006$), and
\textbf{(C)} higher learning experience ratings than \textcolor{BPColorscale}{Blog} ($p=0.026$) and \textcolor{VDColorscale}{Video} ($p=0.033$), on follow-up pairwise comparisons after Kruskal–Wallis tests.
  }
  \label{fig:user-study-overall}
  \Description{Figure that summarizes the results of a between-subject user study with 90 participants comparing Transformer Explainer, Blog, and Video on quiz accuracy, self-ratings of learning objectives, and personal experience. The left panel shows average quiz accuracy: Transformer Explainer participants achieved a mean of 73.3\%, significantly higher than Blog at 59.5\% and Video at 61.9\% (p = 0.021). The middle panel presents average self-ratings for achieving learning objectives, with Explainer rated 3.59, significantly higher than Blog at 3.02, while Video was 3.54 with no significant difference. The right panel displays personal experience ratings, where Explainer averaged 3.61, significantly higher than Blog at 2.94 and Video at 3.17. Overall, Transformer Explainer consistently outperformed both baseline materials across these three measures, with statistically significant advantages in accuracy, perceived learning achievement, and subjective experience.}
\end{figure*}

\subsection{Data Analysis}
We employed a mixed-methods approach to analyze the study data, combining quantitative and qualitative techniques to address our research questions. 
The quantitative analyses (\autoref{sec:quant}) examined participants’ objective quiz performance and subjective ratings for achieving the learning objectives (\autoref{tab:lo_quiz}) to evaluate the effectiveness of each learning resource in supporting Transformer concept understanding (\ref{rq:understanding}).
We also asked participants to rate their personal experience using each material (\ref{rq:experience}).
In addition, \textcolor{TEColorscale}{\tool{}} participants provided feature-usefulness ratings, allowing us to investigate how they engaged with the tool and to uncover usage patterns associated with successful learning (\ref{rq:feature}).
To complement these measures, we conducted a qualitative thematic analysis (\autoref{sec:qual}) of participants’ open-ended responses to capture deeper insights into their  experiences, sources of confusion, and suggestions for improvement.

\subsubsection{Quantitative Analysis}
\label{sec:quant}
\paragraph{Quiz Accuracy.}
Quiz accuracy was measured with a 7-question multiple-choice quiz aligned with six learning objectives (\autoref{tab:lo_quiz}). 
We analyzed quiz accuracy using generalized linear mixed models (GLMMs) \cite{breslow1993approximate} with a binomial logit link. 
Condition (\textcolor{TEColorscale}{\tool{}}, \textcolor{BPColorscale}{Blog}, \textcolor{VDColorscale}{Video}) was entered as a fixed effect, and participant was modeled as a random intercept to account for repeated responses per participant.
Participants' self-reported math proficiency and generative AI familiarity were included as covariates to control for prior knowledge, but neither showed reliable effects ($p>0.20$). 
Quiz question was initially included as an additional fixed effect, but since its effect was not significant ($p>0.30$), we report models focusing on condition-level differences. 
For significant condition effects, we conducted planned one-sided contrasts (\textcolor{TEColorscale}{\tool{}} > \textcolor{BPColorscale}{Blog}, \textcolor{TEColorscale}{\tool{}} > \textcolor{VDColorscale}{Video}), 
applying Holm correction across the two comparisons \cite{holm1979simple}.

\paragraph{Subjective Ratings.}
Participants reported their self-perceived understanding of each learning objective (\autoref{tab:lo_quiz}) on a 5-point Likert scale, as well as overall learning experience including usability, engagement, clarity, and self-efficacy. Cognitive load was measured on a 0--10 scale. 
Because the data were ordinal and Shapiro-Wilk tests~\cite{shapiro1965analysis} indicated violations of normality, we used non-parametric Kruskal-Wallis tests~\cite{kruskal1952use} to examine group differences. Planned one-sided contrasts (\textcolor{TEColorscale}{\tool{}} > \textcolor{BPColorscale}{Blog}, \textcolor{TEColorscale}{\tool{}} > \textcolor{VDColorscale}{Video}) were tested with Holm correction across the two comparisons \cite{holm1979simple}. 
\textbf{Usability} was measured with the UMUX-Lite scale for \textit{ease of understanding}, and \textit{usefulness} \cite{lewis2013umux-lite}; 
\textbf{engagement} with measures adapted from the Intrinsic Motivation Inventory for \textit{enjoyment}, \textit{interestingness}, \textit{attention-holding} \cite{lewis2013umux-lite}; 
and \textbf{mental demand} with a measure from NASA-TLX~\cite{hart1988development}. 
We also measured \textbf{clarity}  (\textit{``I could follow the explanations without much confusion''}) and \textbf{self-efficacy} (\textit{``I feel confident that I can explain the basics of a Transformer''}). 
Internal consistency for the two \textit{usability} measures and the three \textit{engagement} measures was high, with Cronbach's $\alpha=0.82$ and $\alpha=0.88$ respectively \cite{cronbach1951coefficient}, supporting the use of averaged scores across measures.

\subsubsection{Qualitative Analysis}
\label{sec:qual}
Participants also responded to three open-ended questions: (1) what aspects of the material were most helpful, (2) what aspects were confusing, and (3) what improvements they would suggest. 
We analyzed these responses using thematic analysis following Braun and Clarke’s six-phase approach \cite{braun2006thematic}. 
We conducted open coding to identify meaningful units of text, then iteratively grouped codes into candidate themes. Through refinement, we developed a final codebook that captured recurring patterns across responses. Representative quotes for each theme were  selected to illustrate the findings.

\begin{figure*}[tb]
  \centering
  \includegraphics[width=0.85\linewidth]{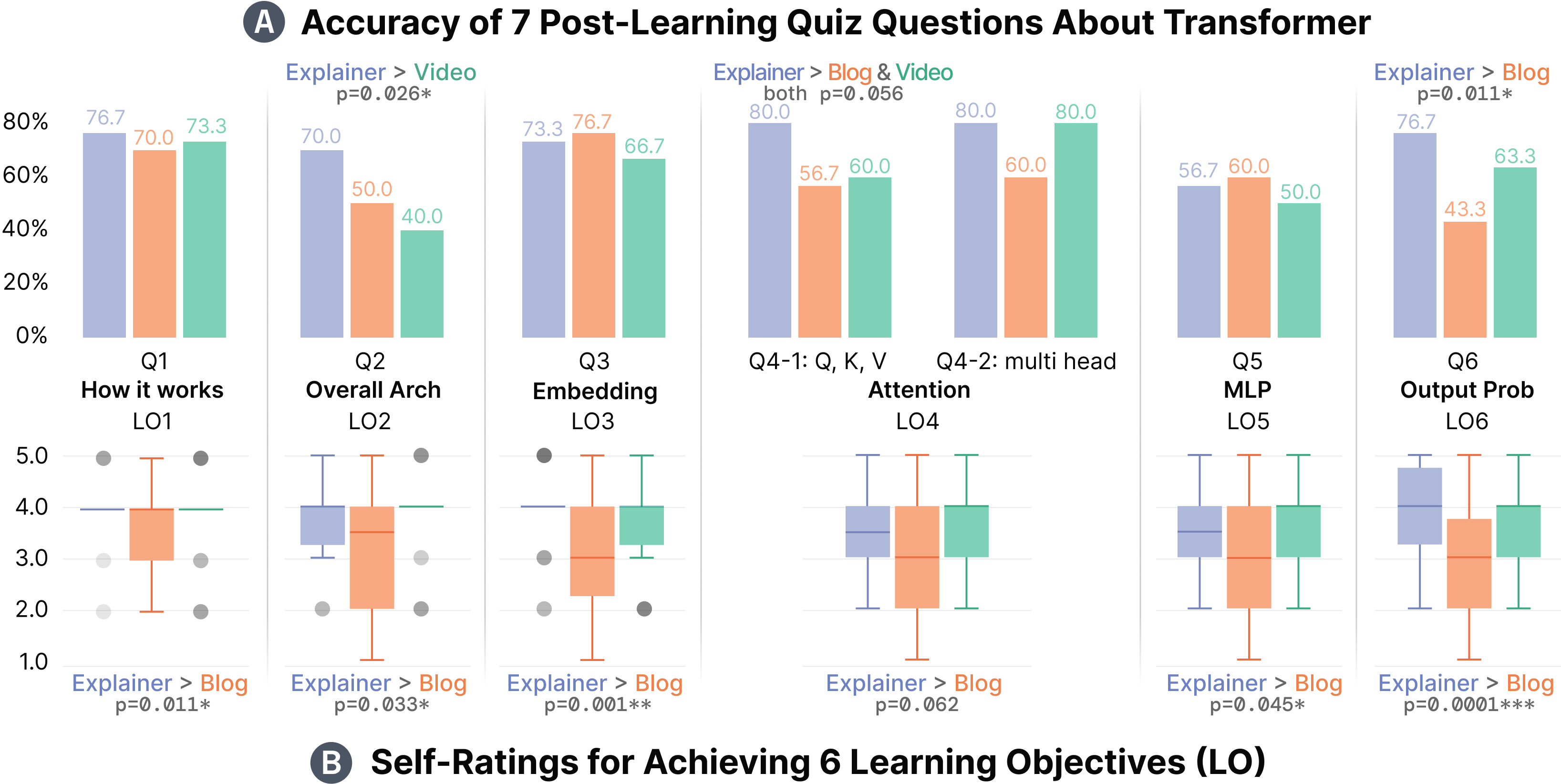}
  \caption{
Across all learning objectives, \textcolor{TEColorscale}{\tool{}} led to significantly higher understanding scores than both \textcolor{BPColorscale}{Blog} and \textcolor{VDColorscale}{Video}.
\textbf{(A)} Across the seven quiz questions aligned with six learning objectives, \textcolor{TEColorscale}{\tool{}} participants scored higher on \textit{architecture overview} (LO2) than \textcolor{VDColorscale}{Video} ($p=0.026$); marginally higher on \textit{attention mechanism} (LO4) than both \textcolor{BPColorscale}{Blog} and \textcolor{VDColorscale}{Video} (both $p=0.056$); and significantly higher on \textit{output probability} (LO6) than \textcolor{BPColorscale}{Blog} ($p=0.011$). 
\textbf{(B)} For self-reported achievement of the learning objectives, \textcolor{TEColorscale}{\tool{}} participants rated themselves significantly higher than those in \textcolor{BPColorscale}{Blog} across all items.
  }
  \label{fig:quiz-understanding}
  \Description{Figure that presents participants’ performance on seven post-learning quiz questions about Transformers and their self-ratings for achieving six corresponding learning objectives. The top row shows bar charts of quiz accuracy, where Transformer Explainer generally outperforms Blog and Video. On Question 1 about “How it works,” Explainer accuracy is 76.7\%, higher than Video at 73.3\% and Blog at 70.0, with a significant advantage over Video. On Question 2 about overall architecture, Explainer reached 70.0\%, clearly higher than Blog at 50.0\% and Video at 40.0\%. On Question 3 about embeddings, Explainer scored 73.3\%, above Blog at 70.0\% and Video at 66.7\%. On Question 4 about attention, scores were closer, with Explainer at 56.7\% and Blog at 60.0\% for Q, K, V, while Explainer and Blog both reached 80.0\% for multi-head attention. On Question 5 about MLP, Explainer scored 56.7\%, slightly below Blog at 60.0\% but above Video at 50.0\%. On Question 6 about output probabilities, Explainer achieved 76.7\%, substantially higher than Blog at 43.3\% and Video at 63.3\%, showing a significant difference. The bottom row shows boxplots of self-ratings for achieving learning objectives, again with Explainer consistently rated higher than Blog across objectives. Significant differences are observed for LO1 (p = 0.011), LO2 (p = 0.033), LO3 (p = 0.001), LO5 (p = 0.045), and LO6 (p < 0.001). Overall, the results indicate that Transformer Explainer not only improved quiz accuracy on several Transformer concepts but also increased participants’ perceived success in meeting key learning objectives compared to the baselines.}
\end{figure*}

\section{Findings and Reflections}
\label{sec:findings}

Here, we present the findings for our three research questions:
% described in \autoref{sec:user_study}: 
\autoref{sec:understanding} discusses how our tool improves non-expert learners' understanding (\ref{rq:understanding});
\autoref{sec:experience} describes how learners' personal experience is enhanced (\ref{rq:experience});
and \autoref{sec:finding-rq3} discusses how our tool's specific features support effective learning (\ref{rq:feature}).
Finally, \autoref{sec:lessons} reflected on our lessons learned on user needs, explanation effectiveness, and learning outcomes.
%
% integrated analysis that combines objective performance, self-reported understanding, and participants’ reflections to show not only learning outcomes, but how specific design features contributed to those outcomes.

\subsection{How does \tool{} improve non-expert learners’ understanding of Transformer concepts compared to blog posts and videos, which are popular means of learning? (\ref{rq:understanding})}
\label{sec:understanding}

\subsubsection{Summary}
Our analyses present converging evidence that our tool improves non-experts’ understanding of Transformer concepts more effectively than  \textcolor{BPColorscale}{Blog} and \textcolor{VDColorscale}{Video} (\autoref{fig:user-study-overall}), showing \textbf{statistically significant} advantages in both \textit{objective} quiz accuracy and \textit{subjective} ratings in achieving learning objectives.  
\autoref{fig:user-study-overall}A shows \textcolor{TEColorscale}{\tool{}} participants answered statistically significantly more quiz questions correctly ($73.3\%$ correct on average across 7 questions) than those in \textcolor{BPColorscale}{Blog} ($p=0.021$) and \textcolor{VDColorscale}{Video} ($p=0.021$).
They also self-rated their understanding as significantly higher 
than \textcolor{BPColorscale}{Blog} ($p=0.006$), and comparable to \textcolor{VDColorscale}{Video} (\autoref{fig:user-study-overall}B).

\subsubsection{Question and learning objective level analysis}
\textcolor{TEColorscale}{\tool{}}'s benefits are further highlighted by analyses at the quiz question and learning objective level. 
\autoref{fig:quiz-understanding}A shows that participants using \textcolor{TEColorscale}{\tool{}} achieved significantly higher quiz accuracy than \textcolor{BPColorscale}{Blog} on the \textit{output probability} (LO6), and marginally higher than both \textcolor{BPColorscale}{Blog} and \textcolor{VDColorscale}{Video} on the \textit{attention mechanism} (LO4)---concepts often regarded as the main reasons of difficulty when learning Transformers. 
They also outperformed \textcolor{VDColorscale}{Video} on the \textit{architecture overview} (LO2), a topic that requires a clear understanding of the model’s overall structure.

\textcolor{TEColorscale}{\tool{}} participants self-rated significantly higher in achieving all learning objectives (\autoref{fig:quiz-understanding}B) compared to \textcolor{BPColorscale}{Blog}, except for the \textit{attention mechanism} (LO4) where the difference was marginally significant.
While participants in the \textcolor{VDColorscale}{Video} condition reported understanding levels comparable to \textcolor{TEColorscale}{\tool{}}, their overall quiz accuracy, averaged across questions, was as low as \textcolor{BPColorscale}{Blog}'s, and significantly lower than \textcolor{TEColorscale}{\tool{}}'s (\autoref{fig:user-study-overall}A).
This pattern may reflect an “illusory understanding” effect \cite{rozenblit2002misunderstood},
where prior research suggests that watching a seemingly coherent narrative (e.g., a video) can create the  impression of understanding without supporting  recall \cite{deslauriers2019measuring}.
Indeed, one \textcolor{VDColorscale}{Video} participant rated their understanding 4 out of 5, but admitted: \textit{``When I got to the quiz, I didn’t really remember the details. I think maybe short quizzes along the way [would help]''} 
Another noted, \textit{``it was not confusing, but hard to retain. I am a visual person and remembering without the video is difficult for me.''}
Several others ($n=3$) explicitly suggested adding quizzes or interactive activities during video viewing to improve retention.

Interestingly, \textcolor{TEColorscale}{\tool{}} participants commented on how the tools provide the type of interactive elements that are ``missing'' from \textcolor{VDColorscale}{Video}.
One participant explained, \textit{``I like how it is interactive and allowed me to change certain settings. I find interactive learning helps me better understand and retain compared to only reading.''}
Another noted, \textit{``I really found the interactive visuals useful in aiding comprehension while I was doing the quiz.''}

\begin{figure*}[!t]
  \centering
  \includegraphics[width=0.85\linewidth]{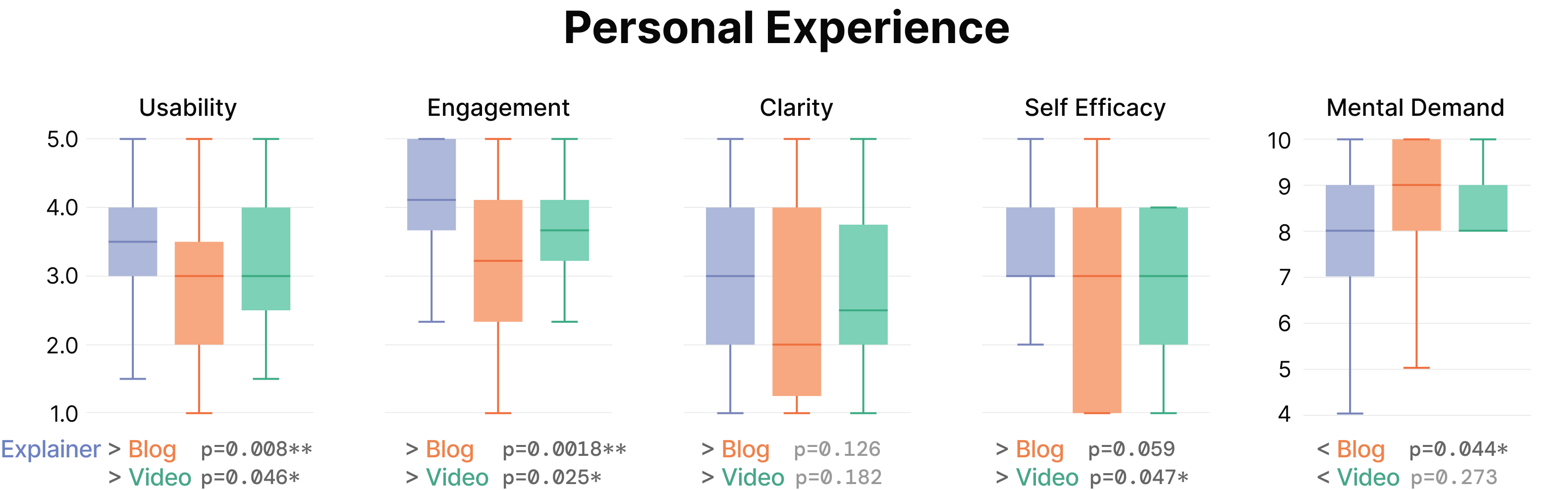}
  \caption{
For participants' personal experience using the three tools, \textcolor{TEColorscale}{\tool{}} outperformed both \textcolor{BPColorscale}{Blog} and \textcolor{VDColorscale}{Video} on all measures. 
Usability and engagement were significantly higher than both baselines, self-efficacy was significantly higher than \textcolor{VDColorscale}{Video} ($p=0.047$), and mental demand was substantially lower than \textcolor{BPColorscale}{Blog} ($p=0.044$). 
  }
  \label{fig:usability}
  \Description{Figure that shows five boxplots comparing participants’ ratings of Transformer Explainer, Blog, and Video across usability, engagement, clarity, self-efficacy, and mental demand. Ratings use a 1–5 scale for the first four measures and a 1–10 scale for mental demand. Transformer Explainer received significantly higher scores than Blog and Video on usability and engagement, showed slightly higher clarity ratings without significant differences, and achieved higher self-efficacy, significantly above Video. For mental demand, Transformer Explainer was rated lower than Blog, indicating reduced cognitive load, while the difference from Video was not significant. Overall, Transformer Explainer outperformed both baseline materials on most measures.}
\end{figure*}

\subsection{How is learners’ personalized experience enhanced when learning Transformer concepts? (\ref{rq:experience})}
\label{sec:experience}
\autoref{fig:user-study-overall}C shows that 
\textcolor{TEColorscale}{\tool{}} participants rated their overall personal experience 
statistically significantly higher than both  \textcolor{BPColorscale}{Blog} ($p=0.026$) and \textcolor{VDColorscale}{Video} ($p=0.033$).
Breaking this down by measures (\autoref{fig:usability}),  \textcolor{TEColorscale}{\tool{}} consistently outperformed the baselines across multiple measures.
 Specifically, on \textbf{usability} (\textit{ease of understanding}, \textit{usefulness}) and \textbf{engagement} (\textit{enjoyment}, \textit{interestingness}, \textit{attention-holding}), our tool scored significantly higher than both \textcolor{BPColorscale}{Blog} and \textcolor{VDColorscale}{Video}.
 \textbf{Self-efficacy} ratings were also significantly higher than \textcolor{VDColorscale}{Video} ($p=0.047$), suggesting that learners felt more confident in explaining Transformer basics after using \textcolor{TEColorscale}{\tool{}}.
 As one participant noted: \textit{``I feel that I could map out the process [...] the equivalent of a children’s drawing of the solar system in crayon. I understand that everything moves about one another.''}
On \textbf{mental demand}, \textcolor{TEColorscale}{\tool{}} was rated statistically significantly lower than \textcolor{BPColorscale}{Blog} ($p=0.044$), while being comparable to \textcolor{VDColorscale}{Video}. 
This result is noteworthy because interactive tools are sometimes associated with higher cognitive load due to user actions \cite{skulmowski2022understanding, song2014cognitive}, whereas videos are typically regarded as cognitively lightweight since learners only need to follow the narrative \cite{fan2024video}.
We attribute this outcome to two design features: a flow-based visualization with visual consistency (\ref{goal:flow}), and step-by-step guided explanations (\ref{goal:math}) that scaffolded learners through abstraction levels appropriate for non-experts (\autoref{sec:finding-rq3}).

For clarity, all three tools' scores were modest, reflecting the inherent challenge of the Transformer topic for non-experts with little prior background knowledge. 
Nonetheless, \textcolor{TEColorscale}{\tool{}} participants experienced the highest average clarity (at 3.07), 
while reported the best ratings across all other measures. 
Taken together, these findings demonstrate the viability of interactive visualization approaches in enhancing participants' personal experience when learning Transformer concepts.

% \subsubsection{Lesson Learned}
% Although interactive tools are sometimes associated with higher cognitive demand, our results suggest that this burden can be substantially reduced when interactions are supported by two design principles. 
% % 
% First, the token-centric flow visualization (\ref{goal:flow}) helps learners build a clear visual image of how their own input data is transformed, which in turn increases self-efficacy and reduces mental demand. 
% % 
% Second, when introducing particularly complex concepts, our visual disambiguation and step-by-step guided explanations (\ref{goal:math}) allow non-experts to avoid feeling overwhelmed by first gaining a visual intuition and then exploring abstraction levels at their own pace.
% % 
% Taken together, these findings indicate that appropriately scaffolded interactivity and token-centric flow visualization can enhance the tool’s usefulness while avoiding common pitfalls of cognitive overload.

\subsection{How do the features of \tool{} support effective learning? (\ref{rq:feature})}
\label{sec:finding-rq3}
Our results suggest that the effectiveness of \textcolor{TEColorscale}{\tool{}} may be attributed to its success in achieving three design goals that we identified (\autoref{sec:design_goals}): 
enabling dynamic experimentation with live models through user input and hyperparameter manipulation (\ref{goal:experiment}), 
maintaining coherence with flow-based visualizations (\ref{goal:flow}), 
and providing step-by-step guided explanation (\ref{goal:math}). 
Each of these features contributed to higher engagement and lower mental demand, and together they led \textcolor{TEColorscale}{\tool{}} to achieve stronger overall usability evaluations (\autoref{fig:user-study-overall}C).

\subsubsection{Interactivity Enhances Understanding of Transformer Concepts.}  
Unlike the \textcolor{BPColorscale}{Blog} and \textcolor{VDColorscale}{Video}, only \textcolor{TEColorscale}{\tool{}} allowed participants to directly manipulate inputs and model hyperparameters while observing immediate system feedback (\ref{goal:experiment}, \autoref{sec:generation}, \autoref{sec:sampling}). 
In addition, the step-by-step expanded view enabled learners to transition between overview and detail at their own learning pace (\ref{goal:math}, \autoref{sec:expandedView}). 
These features not only contributed to an enhanced personal experience (\autoref{fig:usability}) but were also strongly associated with improved participants' understanding. 
In open-ended responses, 26 of 30 participants \textbf{explicitly} identified interactivity as \textbf{one of the most helpful features} for achieving the learning objectives. Many specifically highlighted entering their own text and adjusting sampling parameters as particularly impactful.
One participant noted,
\textit{``The most helpful part to learn is the interactiveness with the generate button, being able to add my own words to play around, as well as the probabilities of the words coming next being able to play with the sample parameters.''} 
Another remarked,
\textit{``The material that helped me the most was the interaction piece, looking at how temperature, p and k change with the different words and how much creativity I want it to have.''}

\subsubsection{Token-centric Flow-based Visualization Clarifies Complex Model Architecture.}
Although all three learning materials relied heavily on visual explanations, the flow-based visual design of \textcolor{TEColorscale}{\tool{}} offered participants a clearer and more coherent experience.
As shown in \autoref{fig:features}, flow visualization was rated as the second most helpful feature, and in open-ended responses, participants praised its clarity in mapping data trajectories onto the actual model structure ($n=12$). One noted,
\textit{“I really liked seeing the flow visuals of the transformer. I finally have a clear mental image for how it all works.”}
and another,
\textit{“The graphic was very efficient, and made me understand more clearly what was happening at each step.”}
Such comments highlight how this design translated abstract operations into tractable narratives, helping users build a clearer mental model of the Transformer’s internal architecture.

In contrast, \textcolor{BPColorscale}{Blog} and \textcolor{VDColorscale}{Video} participants described visual overload. One participant commented, 
\textit{“The visuals were somewhat helpful, but they could be confusing as well. Too much information at once, including technical terms and graphics.”}
while another noted,
\textit{“Have it be less columns and charts. My eyes just glaze over and attention drifts after the first ten to fifteen minutes of that.”} 
This suggests that \textcolor{TEColorscale}{\tool{}} reduced participants’ mental demand by maintaining a consistent Sankey-style flow across components and integrating overview and detail within a single framing (\ref{goal:flow}, \autoref{sec:overview}), thereby avoiding the introduction of new charts or diagrams for each subtopic as in the \textcolor{BPColorscale}{Blog} and \textcolor{VDColorscale}{Video}—as reflected in the lower mental demand scores compared to the baselines (\autoref{fig:usability}). 

These findings demonstrate the importance of a consistent visual narrative that spans the entire learning process. 
Rather than offering many disparate visual elements, representing complex concepts through a unified visual language reduces learners’ cognitive burden and facilitates deeper understanding. 

\subsubsection{Guided Learning Reduces Entry Barriers and Improves Clarity.}  
\begin{figure}[t]
  \centering
  \includegraphics[width=\linewidth]{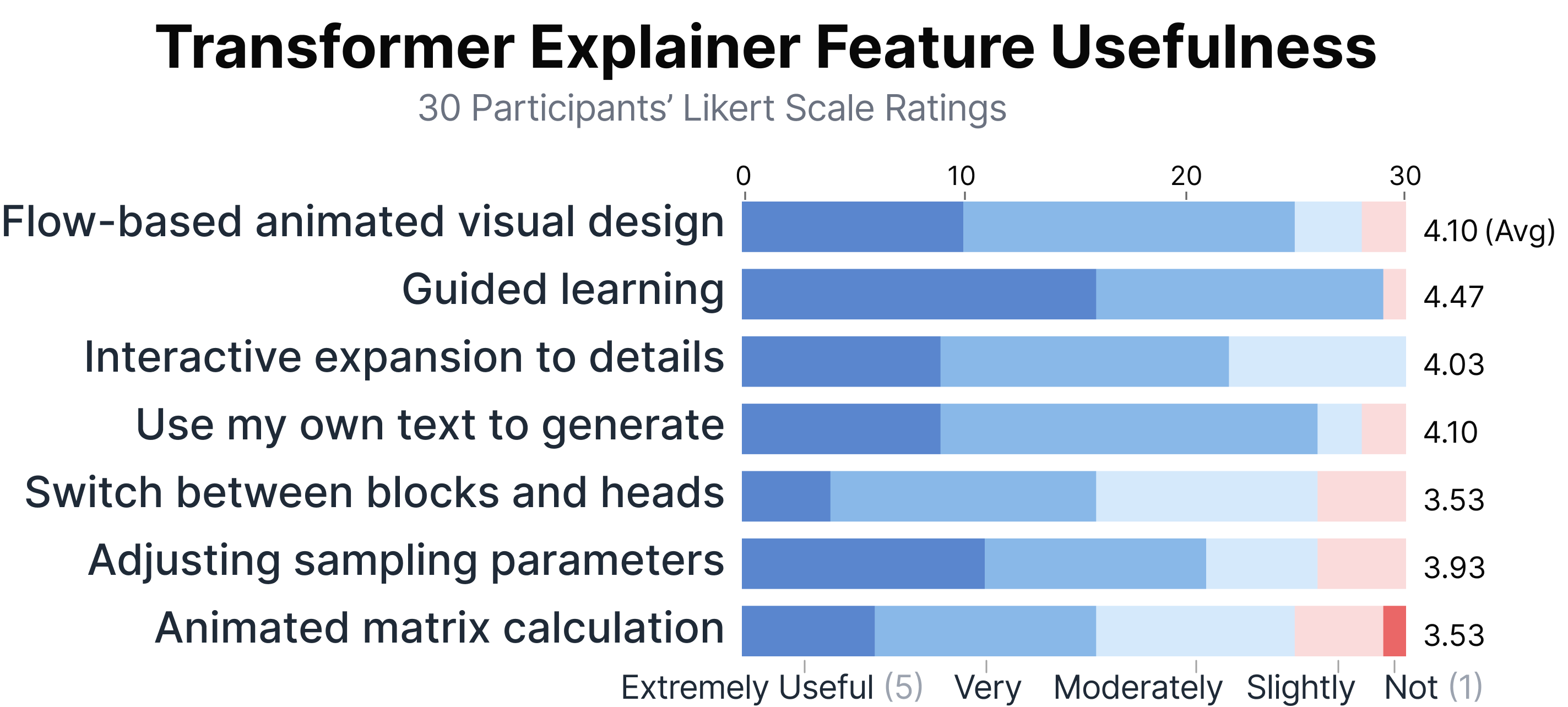}
  \caption{
     Participants rated most features of \textcolor{TEColorscale}{\tool{}} as useful for learning. 
     The highest ratings were given to \emph{guided learning}, \emph{flow-based animated visual design}, and \emph{use my own text}, with a majority of participants rating them as ``extremely useful” or ``very useful.” 
     Other features also received generally positive responses.
  }
  \Description{Figure that shows Transformer Explainer Feature Usefulness. Horizontal bar chart showing usefulness ratings of seven tool features by 30 participants on a 5-point Likert scale (1 = Not useful, 5 = Extremely useful). The highest-rated feature is Guided learning (average 4.47), followed by Flow-based animated visual design (4.10), Use my own text to generate (4.10), and Interactive expansion to details (4.03). Lower-rated features include Adjusting sampling parameters (3.93), Switch between blocks and heads (3.53), and Animated matrix calculation (3.53). Bars are color-coded by response category from dark blue (Extremely useful) to red (Not useful).}
  \label{fig:features}
\end{figure}

As shown in \autoref{fig:features}, the guided learning feature of \textcolor{TEColorscale}{\tool{}} (\autoref{sec:guided-learning}) was rated the highest among all features, as “extremely” or “very” helpful by most participants. 
Several participants emphasized how the structured walkthrough supported their learning.
One noted,
\textit{“I relied a lot on the guided tour so that I could track the progress from start to finish. The interactive design was pretty effective with keeping things moving left→right, and the 1–20 stepped process was essential for tracking over the loops.”} 
and another,
\textit{“The part that helped most was the box on the bottom right corner. It broke down each concept one by one and had helpful navigation on the site.”}  
These scaffolds directly addressed issues observed in the preliminary usability assessment (\autoref{sec:phase2}), where some participants reported not knowing ``what to look at first.” 
Now,
such comments have virtually disappeared.

\subsection{Reflections on User Needs, Explanation Effectiveness, and Learning Outcomes}
\label{sec:lessons}

\subsubsection{[User Needs]
\textcolor{black}{\tool{}} promotes active learning by combining visual narrative with interactive exploration.} 

Reflecting on our design process and study results, we identify several needs of non-expert learners trying to understand Transformers: 
the need to work with personally meaningful inputs, to experiment with model behavior in a low-stakes way, and to receive immediate feedback that makes abstract mechanisms concrete. 
Active learning research suggests that when learners work with inputs they have personally chosen, they develop a stronger sense of stake in the outcome, leading the brain to treat the information as more valuable and thus increasing their readiness to learn and retain \cite{freeman2014active, deslauriers2019measuring}. 
\textcolor{TEColorscale}{\tool{}} directly supports these needs by allowing participants to
(1) choose their own prompts and settings and immediately observe the resulting attention patterns;
(2) interactively modify temperature and sampling hyperparameters while seeing how such changes alter the next-token distribution.
This tight experiment–feedback loop turns parameters that are often discussed only in the abstract into manipulable objects, in ways that passive media such as \textcolor{VDColorscale}{Video} or \textcolor{BPColorscale}{Blog} cannot easily provide.

\textcolor{TEColorscale}{\tool{}}'s higher quiz accuracy (\autoref{sec:understanding}) on the more challenging topics such as \textit{attention mechanism} (LO4) and \textit{output probability} (LO6), both of which supported by rich interaction, suggests that addressing these needs for hands-on experimentation and immediate feedback can translate into better understanding of model internals.
At the same time, participants in the \textcolor{VDColorscale}{Video} condition frequently reported difficulty retaining material and expressed a  desire for  additional mid-lesson activities to reinforce key concepts, underscoring that learners may need more than continuous exposition as they learn. \tool{} addresses this need in part through interactive exploration and guided learning, and future work could explore adding more structured practice activities.
Overall, our findings highlight the value of tools that blend visual explanations with opportunities for active engagement and experimentation, and point to user needs (e.g., personal relevance, controllable pacing, and concrete feedback) that future AI education tools should explicitly support.

\subsubsection{[Explanation Effectiveness] \textcolor{black}{\tool{}} eases non-experts' interpretation of complex Transformer mechanisms.}
Participants using our tool rated \textbf{usability} (ease of understanding and perceived usefulness) significantly higher than \textcolor{BPColorscale}{Blog} and \textcolor{VDColorscale}{Video}, 
and reported higher \textbf{self-efficacy}, indicating greater confidence in their ability to explain core Transformer mechanisms after interacting with the system. 
At the same time, \textbf{mental demand} was rated lower than in both baselines, counter to the common assumption that interactivity necessarily increases cognitive load. This finding suggests that our interactive explanations helped participants make sense of the model's internal processes without feeling overwhelmed, rather than burdening participants with interface complexity. 
Taken together, these three self-reported measures indicate that participants experienced the explanations as both \textit{accessible} and \textit{empowering}. This subjective picture is echoed by the objective quantitative results (\autoref{sec:understanding}).
Specifically, quiz accuracy averaged 73.3\% across seven learning-objective-aligned questions (\autoref{fig:user-study-overall}), a statistically significant improvement over \textcolor{BPColorscale}{Blog} and \textcolor{VDColorscale}{Video}. 
In other words, participants did not only feel that the explanations were clear; they were also better able to answer mechanism-focused questions correctly.
In the next section, we examine these learning outcomes in more detail and discuss how they may be attributed to specific design choices in our tool's explanatory features. 
% 
% This demonstrates that our tool not only felt easier to interpret, but measurably improved learning outcomes. 
% 

\subsubsection{[Learning Outcomes] From Design Features to Learning Gains}

Our evaluation suggests that \textcolor{TEColorscale}{\tool{}} not only improves overall quiz performance and self-rated understanding, but does so in a way that varies across learning objectives in informative ways. Across the seven quiz questions aligned with six learning objectives, participants using \textcolor{TEColorscale}{\tool{}} achieved statistically significant gains on \textit{architecture overview} (LO2, $p=0.026$), and \textit{output probabilities} (LO6, $p=0.011$), and a marginally higher performance on \textit{attention mechanism} (LO4, $p=0.056$), outperforming \textcolor{BPColorscale}{Blog} and often \textcolor{VDColorscale}{Video} on these challenging concepts. These patterns indicate that our tool is particularly effective at helping learners construct a coherent mental model of the model's global structure and the probabilistic nature of text generation, areas where non-experts often default to a ``black box'' view of Transformers.

These outcome differences closely track where our design offers the richest interactive support. Architecture overview (LO2) is reinforced by the flow-based visualization and block navigation (\autoref{sec:overview}), which let learners move between overview and detail on demand; attention mechanism (LO4) is supported through head-level navigation (\autoref{sec:overview}) and step-by-step animated explanations (\autoref{sec:expandedView}); and output probabilities (LO6) are made tangible through direct experimentation with temperature, top-k, and top-p (\autoref{sec:sampling}). In contrast, MLP (LO5)---the only objective without dedicated interactive elements---shows comparatively lower quiz accuracy and self-rated understanding (\autoref{fig:quiz-understanding}). This divergence suggests  our interactive visual scaffolds are not merely engaging add-ons but are tightly coupled to conceptual learning, and it highlights MLP as a concrete target for future work (e.g., by adding analogous visual and experimental affordances).

Finally, comparing subjective and objective outcomes reveals how different resource formats shape learners' sense of understanding. As noted in \autoref{sec:understanding}, participants in the \textcolor{VDColorscale}{Video} condition reported understanding comparable to \textcolor{TEColorscale}{\tool{}}, despite substantially lower quiz accuracy. Prior work characterizes this kind of divergence as possible ``illusory understanding'' effect, where coherent narrative can foster confidence without supporting recall \cite{deslauriers2019measuring}. Building on this, our results suggest that \textcolor{TEColorscale}{\tool{}}'s interactive manipulation of inputs and parameters may help learners calibrate their understanding more accurately: by actively testing how changes in inputs, attention, or sampling parameters affect model behavior, participants receive immediate feedback about what they do and do not yet understand. For a learning-outcomes perspective, this calibration is itself valuable, even though our current measures focus on short-term conceptual gains rather than long-term retention or transfer.

\section{Discussion, Limitations, and Future Work}
\label{sec:discussion}
Our user studies provide promising evidence of the effectiveness of interactive visualizations for helping non-expert learners understand Transformer models. At the same time, our evaluations surfaced important challenges and opportunities for further refinement. In this section, we discuss current limitations of our tool and outline future directions for broadening and deepening interactive visual explanations for AI education.

\subsection{Extending Support for Other \tf{} Modalities} %
 While \tool{} currently focuses on text-based \tfs{}, we observed from our user study that participants showed significant interest in seeing the tool expanded to domains beyond language (e.g, \textit{``I would like to learn about how the model behaves when its used for generating images. How does it differ from this one?''}).
 Many participants expressed curiosity about how the same architectural principles manifest in other domains, particularly multimodal and non-linguistic applications.
 Extending support to models such as Vision Transformers (ViT) for vision \cite{vit}, Whisper for speech \cite{whisper}, or vision-and-language models like CLIP and SigLIP \cite{clip, siglip} would highlight that Transformers are not limited to text, but operate as a general-purpose architecture across diverse modalities \cite{xu2024survey}. 
 Our flow-based visual design (\autoref{sec:overview}) and step-by-step explanations (\autoref{sec:expandedView}) could naturally generalize to these architectures, though they would also introduce new challenges. 
 For instance, tasks such as image generation may require specialized visualization techniques from prior work on diffusion models and visual interpretability \cite{lee2024diffusion, ma2023visualizing}, including dimensionality-reduced visual summaries of generation trajectories or saliency-based overlays for visual attention.
 Exploring how these methods might be integrated into our framework presents an exciting direction for future development.

At the same time, expanding the conceptual framing beyond the current \emph{word} = \emph{token} representation offers another avenue for broadening the tool’s reach. 
Our current walkthrough primarily employs text-generative Transformers---the domain most familiar to general audiences—to explain tokenization and embedding using user-provided sentences (\autoref{sec:generation}). 
However, Transformers fundamentally operate on discrete token representations, which need not correspond to linguistic words. 
Protein sequences, gameplay logs, and other forms of structured data can all be represented as tokens and processed in the same way as text. 
Making this abstraction explicit, and offering interactive examples that illustrate how non-linguistic token streams are embedded, transformed, and attended to, could help learners more deeply appreciate the universality of the architecture.

Taken together, these directions suggest that \tool{} could evolve into a cross-domain explainer of Transformer architectures. 
By combining modality-specific visualization techniques with a token-centric framing, the tool could transcend its role as a text-model explainer and become a platform that communicates the generality of Transformers across a wide range of applications. 
Such an expansion would not only benefit researchers and practitioners in various domains, but also help non-experts recognize that systems capable of generating text, interpreting images, recognizing speech, or even modeling biological sequences all rely on the same architectural building blocks. We see this as an opportunity to move from ``explaining GPT-2” toward ``explaining the Transformer paradigm” more holistically.

\subsection{Deepening and Scaling Up Transformer Explanations}
In our study, participants expressed interest not only in exploring larger models and extended contexts (e.g., GPT-4), but also in obtaining deeper insights into core mechanisms like attention.
Supporting these interests requires tackling complementary challenges: enriching interpretive depth while simultaneously scaling visualization strategies for larger and more complex models.

\subsubsection{Deepening Attention Interpretation and Visualization}
Non-expert learners in particular expressed strong interest in understanding the meaning and role of attention, indicating that deeper interpretation could support more comprehensive conceptual understanding. 
As one participant noted, \textit{``I’d love to see how multiple heads collaborate in making predictions. Do some heads agree on key words, or do they specialize and work independently?''} Designing visualizations that make the functions of multi-head attention clearer—how different heads capture varying complexities or attend to distinct aspects of the input—could enhance learning outcomes. 
A fruitful direction is to reflect that attention has no privileged basis in query–key (or value–projection) space \cite{ameisen2025circuit} by finding a rotation that best groups high-activation neurons, making visual patterns easier to interpret.

While existing attention visualization tools \cite{wang2021dodrio, yeh2023attentionviz, li2023does, vig2019bertviz} have largely been designed for researchers, future works could carefully adapt such ideas for non-experts, enabling users to interactively explore head relationships and block-level behaviors while preserving ease of use. Beyond attention, additional opportunities lie in visualizing advanced behaviors such as in-context learning, showing how prepending a few examples changes attention patterns and alters predictions on a new task. 
These expansions could help learners move beyond introductory explanations and explore both the fundamental mechanisms and higher-level capabilities of modern Transformers.

\subsubsection{Supporting Larger Models and Extended Contexts}
Recent \tf{} models increasingly handle longer inputs and richer contextual information. Our user study echoed this interest in educational tools for exploring large-scale models, such as GPT-4. 
However, a long input prompt with many tokens could quickly increase visual complexity, even at the overview abstraction level.
Additionally, running such large models directly in web browsers
remains technically challenging \cite{lee2024diffusion}. 
Future research may therefore explore specialized visualization strategies alongside efficient browser-based implementations  (e.g., WebAssembly optimization \cite{ruan2024webllm}, model compression), 
making large-scale \tf{} models accessible to broader audiences.

\subsection{Limitations of Study Design}
While our study highlights the potential of interactive visualization tools for teaching complex AI concepts, it also carries several  limitations. 
First, we conducted a one-hour user study and capped participants’ tool usage at 45 minutes (\autoref{sec:study-design}). 
This controlled environment allowed fair comparisons across conditions, but may not fully reflect the most natural learning contexts for all participants. 
In practice, learners might spend more than an hour exploring, revisit concepts after a break, or adopt other study rhythms. Such interaction patterns may only emerge over extended periods of use.

Second, our evaluation measured learning outcomes immediately after the session using quizzes and surveys (\autoref{sec:procedure}). 
While this design captures short-term comprehension, it does not assess long-term retention or transfer. 
In this study, we intentionally focused on whether the tool improves non-expert learners’ understanding of high-level Transformer concepts, and therefore excluded detailed assessment of mathematical mechanisms. 
However, such fine-grained understanding may only be measurable after learners have had time to consolidate and re-apply the knowledge. 
Future research may investigate longitudinal effects, evaluating retention and application weeks or even months later.

Third, our participant pool primarily consisted of non-expert learners with limited prior exposure to Transformers (\autoref{sec:participants}). 
This choice was appropriate for studying how the tool supports beginners who are interested in AI but lack formal ML expertise. 
Nonetheless, perspectives from advanced learners, instructors, and practitioners remain important for future work. 
Experts and educators might use the tool differently---for example, as a teaching aid, a debugging resource, or a way to illustrate abstract concepts in the classroom. 
Their feedback would provide valuable insights into how the tool can extend to broader educational contexts. 
Moreover, studies with ML students could compare the tool against lecture materials or textbooks to evaluate whether interactive visualization also aids in understanding finer-grained mathematical mechanisms.

Finally, we compared \tool{} against some of the most popular baseline resources---a widely-read blog post and a highly-viewed YouTube video (\autoref{sec:study-design}). Even against these well-known materials, our tool demonstrated clear advantages. 
Still, the variety of existing blogs and videos is much broader. Expanding the range of baselines in future comparisons would help further validate and contextualize our findings.

\subsection{Positioning Our Work in the AI Education Tool Landscape}
We are the first to visualize all blocks and components of a Transformer as a token-centric data flow (\ref{goal:flow}), 
showing that multi-head self-attention, often viewed as the steepest learning hurdle, can be visually unpacked within a unified graphical language (\ref{goal:math}). 
We further address a core Transformer characteristic that many prior educational tools overlook: its autoregressive and probabilistic behavior. 
To do this, we run a live Transformer model directly in the browser, extract intermediate attention values and output probabilities in real time, and integrate them seamlessly into the visualization. 
Coupled with animations that reveal the model’s internal computation flow, our system enables an immediate experiment–feedback loop that responds to user edits of the input and sampling hyperparameters (\ref{goal:experiment}). 
We validate the effectiveness of this approach through a user study, showing meaningful learning benefits for non-experts without an ML background (\autoref{sec:user_study}). 

Beyond the formal study, we observe substantial educational impact in real-world use. 
Since releasing the initial prototype, the tool has been used by over 490,000 users across over 200 countries and adopted as course material in undergraduate and graduate ML-related courses at leading universities worldwide.
Some instructors report appreciating the ability to use the tool with large classes without server constraints, and both instructors and students value being able to run models live and explore how different hyperparameters shape model behavior.
In addition, we open-sourced the system to broaden accessibility and extensibility. As a result, the community has released versions of Transformer Explainer in multiple languages and begun adapting it to additional models. 

Taken together, our work contributes a novel design  that combines token-centric flow visualization with interactive, in-browser model experimentation, and demonstrating both its educational effectiveness and its potential for broad adoption and extension.

\subsection{The Role of Interactive Visualization in Advancing AI Education}
Our study demonstrates that interactive visualization tools can provide significant benefits for AI education (\autoref{sec:findings}). 
By allowing learners to directly engage with model components and observe the effects of their interactions, such tools transform learning from passive knowledge acquisition into active, exploratory engagement.
In designing our tool, we drew inspiration from earlier examples \cite{karpathyConvNetJSMNISTDemo2016, tensorflowplayground, kahng2019gan, CNNExplainer}, which pioneered visual experimentation with machine learning concepts. At the same time, we acknowledge that developing high-quality educational visualizations can require substantial time and effort. For example, our iterative design process (\autoref{sec:iteration}) extended over a year of refinement and evaluation. 

To mitigate these challenges, recent efforts have explored modularizing visualization components to reduce development effort and accelerate the creation of educational tools for emerging AI architectures (e.g., ManimML \cite{helbling2023manimml}). 
We view this approach as a promising direction: by lowering barriers to tool development, the community can expand the availability of interactive learning resources and better align educational tools with the rapid pace of advances in AI models. We are excited to see how the HCI, visualization, and AI education communities will continue contributing to this effort, building on our work to make AI concepts more accessible and understandable to diverse learners.
\section{Conclusion}
\label{sec:conclusion}
We presented \tool{}, an interactive visualization tool aimed at helping non-experts understand a text-generative \tf{} model. Our tool provides a seamless transition between a data flow-based overview visualization and detailed step-by-step explanations. Users can directly interact with a live GPT-2 model in the browser, experimenting with custom input text and hyperparameters. Results from a user study indicate improved understanding in \tfs{} and user engagement.

\begin{acks}
% \section*{Acknowledgments}
This work was supported in part by NSF awards 2403297 and 2502793, the IITP (MSIT, Korea) grant RS-2024-00353131, and gifts from Google, Amazon, Meta, NVIDIA, Avast, Fiddler Labs, Bosch. Alec Helbling is supported by NSF GRFP.
\end{acks}
\bibliographystyle{ACM-Reference-Format}
\bibliography{references}

\appendix

\end{document}